\documentclass[10pt,twocolumn,letterpaper]{article}

\usepackage{cvpr}              

\usepackage{marvosym}
\usepackage{amssymb}
\usepackage{graphicx}
\usepackage{amsmath}
\usepackage{booktabs}
\usepackage{rotating}
\usepackage{multirow}
\usepackage{enumitem}
\usepackage{algorithmicx,algorithm}
\usepackage{enumitem}
\usepackage{pifont}
\usepackage{makeidx}
\usepackage{epstopdf}
\usepackage[noend]{algpseudocode}
\usepackage[x11names]{xcolor}
\usepackage{colortbl}
\usepackage{color}
\usepackage{makecell}

\definecolor{lightblue}{rgb}{0.761,0.894,0.937} 
\definecolor{white}{rgb}{0.965,0.973,0.988} 
\definecolor{lightpurple}{rgb}{0.914,0.870,0.933} 

\usepackage[pagebackref,breaklinks,colorlinks]{hyperref}
\usepackage[capitalize]{cleveref}
\crefname{section}{Sec.}{Secs.}
\Crefname{section}{Section}{Sections}
\Crefname{table}{Table}{Tables}
\crefname{table}{Tab.}{Tabs.}

\colorlet{dark-blue}{blue!70!black}
\hypersetup{
    colorlinks=true,%
    citecolor=dark-blue,%
    filecolor=dark-blue,%
    linkcolor=red,%
    urlcolor=magenta
}

\def\cvprPaperID{777} 
\def\confName{CVPR}
\def\confYear{2025}

\makeatletter
\def\thanks#1{\protected@xdef\@thanks{\@thanks
        \protect\footnotetext{#1}}}
\makeatother

\begin{document}

\title{FisherTune: Fisher-Guided Robust Tuning of Vision Foundation Models for Domain Generalized Segmentation}

\thanks{ 
This work was supported in part by the National Natural Science Foundation of China No. 62271377, the Key Research and Development Program of Shannxi Program No. 2021ZDLGY0106 and No. 2022ZDLGY0112, the Key Scientific Technological Innovation Research Project by Ministry of Education,  the MUR PNRR project FAIR (PE00000013) funded by the NextGenerationEU and the EU Horizon projects ELIAS (No. 101120237) and AI4Trust (No. 101070190).    \\
\quad \quad \textsuperscript{\Letter} Co-corresponding author.}

\author{Dong Zhao$ ^{1}$, Jinlong Li$ ^{2}$, Shuang Wang$^{1}$ \textsuperscript{\Letter}, Mengyao Wu, Qi Zang$ ^{1}$ \textsuperscript{\Letter}, Nicu Sebe$ ^{2}$, Zhun Zhong$ ^{3}$ \\
$ ^{1}$ School of Artificial Intelligence,
Xidian University, Shaanxi, China \\
$ ^{2}$ Department of Information Engineering and Computer Science, University of Trento, Italy \\
$ ^{3}$ School of Computer Science and Information Engineering, Hefei University of Technology, China \\
}
\maketitle

\begin{abstract}
\vspace{-0.2cm}
Vision Foundation Models (VFMs) excel in generalization due to large-scale pretraining, but fine-tuning them for Domain Generalized Semantic Segmentation (DGSS) while maintaining this ability remains a challenge. Existing approaches either selectively fine-tune parameters or freeze the VFMs and update only the adapters, both of which may underutilize the VFMs' full potential in DGSS tasks. We observe that domain-sensitive parameters in VFMs, arising from task and distribution differences, can hinder generalization.
To address this, we propose \textbf{FisherTune}, a robust fine-tuning method guided by the Domain-Related Fisher Information Matrix (DR-FIM). DR-FIM measures parameter sensitivity across tasks and domains, enabling selective updates that preserve generalization and enhance DGSS adaptability.
To stabilize DR-FIM estimation, FisherTune incorporates variational inference, treating parameters as Gaussian-distributed variables and leveraging pretrained priors. Extensive experiments show that FisherTune achieves superior cross-domain segmentation while maintaining generalization, outperforming both selective-parameter and adapter-based methods.
\end{abstract}

\vspace{-0.2cm}
\section{Introduction}
\label{sec:intro}
\vspace{-0.2cm}

Vision Foundation Models (VFMs), such as CLIP \cite{clip}, DINOv2 \cite{dinov2}, and EVA02 \cite{eva02}, have emerged as powerful tools in computer vision, achieving remarkable generalization across diverse downstream tasks, including cross-domain perception\cite{liu2023explicit,zhao2023semantic, Tang_2024_CVPR, scheibenreif2024parameter, zhang2024improving}, few-shot\cite{moor2023foundation, zhou2022conditional, yao2023visual, zang2024ChangeDiff,pu2023dynamic} and zero-shot perception \cite{li2023blip,khattak2023maple, zhang2023prompt,shu2022test, pu2023memorizing}.
Pre-trained on massive data fields, VFMs encapsulate rich visual representations that can be transferred to numerous applications with minor adaptation\cite{yi2024learning, xia2024unsupervised}. 
Despite this, when it comes to Domain Generalized Semantic Segmentation (DGSS), where the goal is to segment unseen domain images without explicit access to their training data, effectively fine-tuning VFMs while preserving their strong generalization capabilities remains an open challenge.

\begin{figure}[t]
    \begin{center}
    \centering 
    \includegraphics[width=0.5\textwidth]{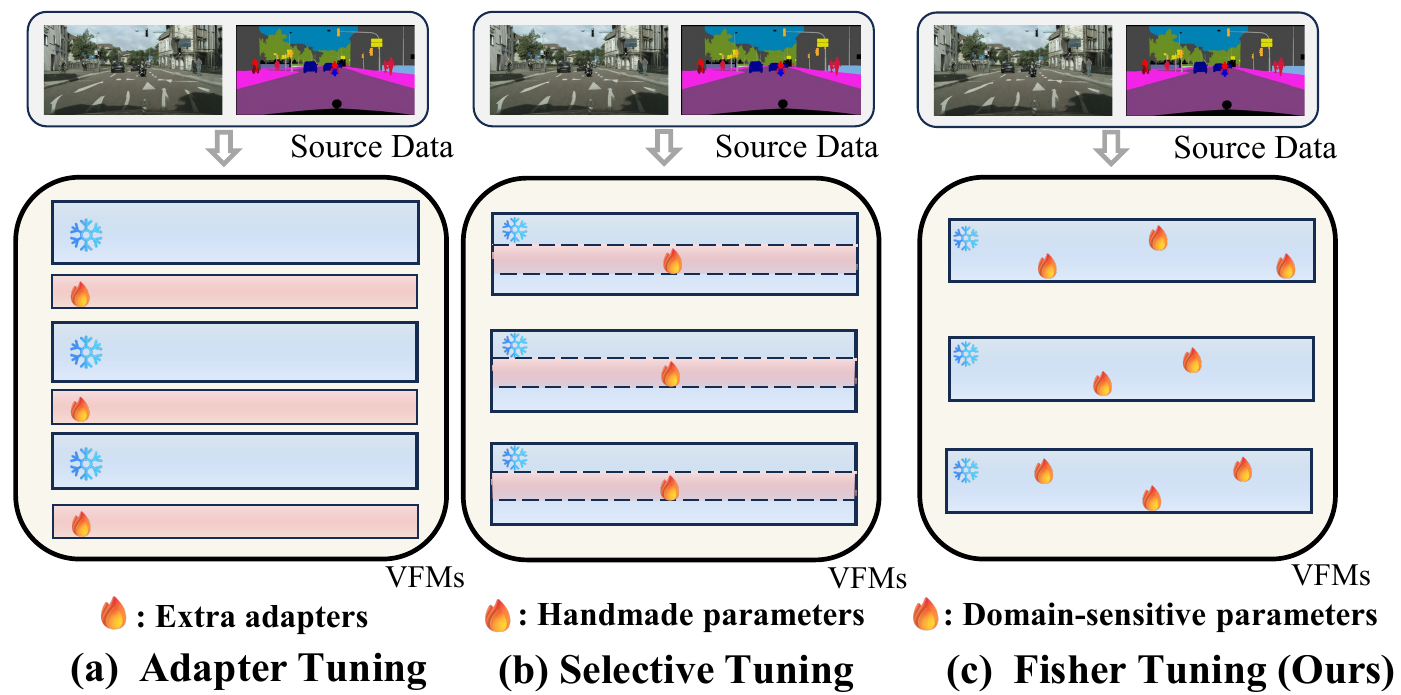} 
    \end{center} 
    \setlength{\abovecaptionskip}{-0.1cm}
    \caption{Comparison of principles for different VFM adjustment methods: (a) tuning by adapter insertion \cite{Wei_2024_CVPR, yi2024learning}, (b) tuning by manually selected \cite{tu2023visual} or automatically selected parameters \cite{sung2021trainingneuralnetworksfixed}, (c) our method for tuning domain-sensitive parameters.}
    \label{eerl}
\vspace{-0.5cm}
\end{figure}

\begin{figure}[h]
    \begin{center}
    \centering 
    \includegraphics[width=0.5\textwidth]{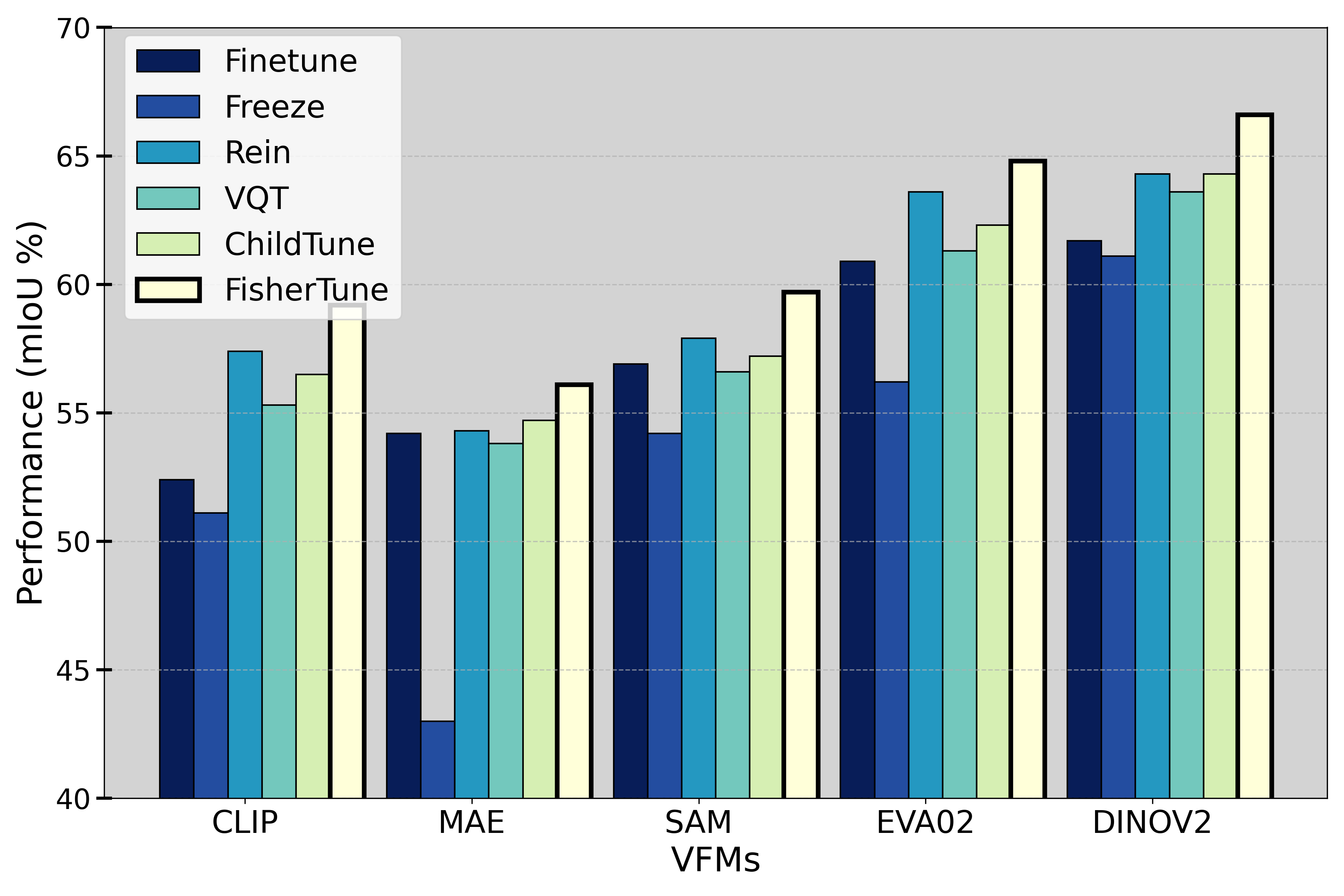} 
    \end{center} 
    \setlength{\abovecaptionskip}{-0.5cm}
    \caption{Comparison of average performance across multiple VFMs in DGSS experiments on GTA → Cityscapes + BDD100K + Mapillary using different fine-tuning methods, including adapter-based Rein\cite{Wei_2024_CVPR}, manually selected parameter-based VQT\cite{tu2023visual}, adaptively selected parameter-based ChildTune\cite{xu2021raisechildlargelanguage}, and our FisherTune.} 
    \label{intro_compare}
\vspace{-0.5cm}
\end{figure}

Existing methods for adapting VFMs to DGSS tasks typically involve fine-tuning via adapter layers to remap pre-trained tokens, as shown in Fig.~\ref{eerl} 
 (a) \cite{Wei_2024_CVPR, yi2024learning}. While this approach reduces overfitting, it does not fully leverage the internal representations of the VFM, as the core content of the model remains unchanged. Furthermore, when the pre-training tasks of the VFM (e.g., MAE, DINOv2) significantly deviate from the DGSS task, the adaptation improvement is limited.
An alternative solution is to fine-tune a subset of parameters related to the target DGSS task, which activates the representations of VFMs, as illustrated in Fig.~\ref{eerl} (b). 
However, we observe that traditional parameter selection methods, whether manually defined \cite{tu2023visual} or automatically chosen \cite{xu2021raisechildlargelanguage, matena2022merging}, fail to guarantee the generalization ability of the VFM, and in fact, perform even worse than simply adding adapters, as shown in Fig.~\ref{intro_compare}.

Our experiments reveal that certain parameters of Vision Foundation Models (VFMs) are crucial for maintaining generalization, while others are key to adapting to new domains and tasks. Traditional selective fine-tuning methods focus solely on task-sensitive parameters, risking the disruption of the VFM’s generalization ability. Therefore, we propose identifying and fine-tuning these domain-sensitive parameters.
To address this, we introduce FisherTune, a novel robust fine-tuning method based on the Domain-Related Fisher Information Matrix (DR-FIM). This method allows us to preserve the generalization capability of the pre-trained VFM while activating its adaptability for DGSS tasks.
Specifically, we first introduce the DR-FIM metric, which measures domain sensitivity by evaluating the fluctuation of parameters across different domains. Unlike FIM metrics, DR-FIM accounts for domain shifts, extending FIM for cross-domain tasks. To mitigate potential degradation in DR-FIM estimation, FisherTune innovatively employs variational inference. By treating model parameters as random variables following a Gaussian distribution and incorporating prior information from the pre-trained VFM, FisherTune stabilizes the DR-FIM estimation process, ensuring accuracy and robustness in cross-domain tasks. This novel estimation method improves computational efficiency and enhances the scalability of FisherTune to large-scale VFM models.
Through extensive experiments on multiple DGSS benchmarks, we demonstrate that FisherTune consistently outperforms both selective-parameter and adapter-based methods.
In summary, our contributions are,
\begin{itemize}
    \item We propose \textit{FisherTune}, a novel fine-tuning strategy that leverages Fisher Information Matrix to selectively fine-tune VFMs for DGSS, preserving the generalization capabilities of VFMs while improving domain adaptability. 
    \item We introduce Domain-Related FIM (DR-FIM), a novel metric matrix that quantifies the sensitivity of parameters to domain shifts.
    \item We employ variational inference to treat model parameters as Gaussian-distributed, ensuring stable and accurate DR-FIM estimation.
    \item We validate the effectiveness of FisherTune through extensive experiments, showing superior generalization compared to state-of-the-art methods.
\end{itemize}

\vspace{-0.1cm}
\section{Related Work}
\vspace{-0.1cm}
\noindent \textbf{Domain Generalized Semantic Segmentation (DGSS)} focuses on enhancing a model's ability to generalize to unseen domains by training on source domains~\cite{ Zhao_2023_CVPR, pan2022fourier, cheng2022domain}. 
Common strategies include domain-invariant representation learning methods and domain augmentation techniques. 
Domain-invariant representation learning approaches involve splitting learned features into domain-invariant \cite{zang_mm_GSF, JSLS_TCSVT} and domain-specific components~\cite{Zhao_2023_ICCV, wang2021learning, li2021uncertainty, 9961139}, or employing meta-learning to develop more robust models~\cite{dou2019domain, li2018learning, Zhao_2024_CVPR_SND}. Additionally, several methods have succeeded by learning feature normalization or whitening schemes~\cite{peng2022semantic, choi2021robustnet, nam2021reducing}. Domain augmentation techniques, on the other hand, improve segmentation results through style transfer at image-level \cite{zhong2022adversarial, peng2021global, pan2022fourier, huang2021fsdr} or feature-level \cite{cheng2022domain, zhao2022style, chattopadhyay2023pasta} and the introduction of additional data~\cite{murez2018image, yue2019domain, lee2022wildnet}.
Some recent work has shown that text-guided feature enhancement \cite{fahes2023poda, fahes2024simple} through CLIP \cite{radford2021learning} or synthetic data of diffusion models can also benefit model generalization\cite{niemeijer2024generalization, benigmim2024collaborating, zhao2023semantic}.


\noindent \textbf{Parameter-Efficient Fine-Tuning}(PEFT) \cite{han2024parameter} customizes pre-trained models by fine-tuning a subset of parameters\cite{gao2024mini}, improving performance and generalization with lower computational cost. The dominance of ViT \cite{dosovitskiy2020image} in vision tasks has spurred the development of PEFT methods. Adaptor-based Prompt Tuning \cite{liao2023parameterefficientfinetuningintroducingnew, sung2021trainingneuralnetworksfixed} has shown strong performance in vision transfer tasks by adding learnable prompts. For instance, Visual Prompt Tuning (VPT) \cite{jia2022visual} introduces learnable prompts for each Transformer layer's input embeddings, AdaptFormer \cite{chen2022adaptformer} adds a bottleneck fully connected layer parallel to the MLP block, and VQT \cite{tu2023visual} optimizes prompts through bypassing to effectively leverage intermediate features of VFMs.

In semantic segmentation, several works have applied visual prompts for model transfer. \cite{liu2023explicit} uses frequency and spatial prompts to transfer pre-trained ViTs to low-level segmentation tasks, while \cite{yang2024exploring} applies mask prompts to aid continual adaptation. \cite{zhang2024improving} uses weak supervision for LoRA adapter \cite{lora} to adapt VFMs across domains. 
Rein \cite{Wei_2024_CVPR} adds LoRA adapters to transform tokens and activate VFMs for DGSS. \cite{yi2024learning} enhances cross-domain adaptation by adding Fourier transform prompts to intermediate tokens of VFMs.

While most visual semantic segmentation methods rely on adapter-based prompt fine-tuning, our work pioneers selective fine-tuning methods for visual model adaptation. Closely related to our approach are selective fine-tuning methods in NLP, like ChildTune \cite{xu2021raisechildlargelanguage} and Fishermask \cite{matena2022merging}, which use Fisher matrices to identify task-sensitive parameters. In contrast, our method introduces domain-sensitive parameters and a stable estimation method to identify parameters highly sensitive to both tasks and domains, making it particularly suited for DGSS tasks.

\vspace{-0.1cm}
\section{Methodology}
\label{sec:method}
\vspace{-0.1cm}

\subsection{Preliminaries}


\noindent \textbf{Domain Generalized Semantic Segmentation} (DGSS) aims to train models that can generalize across unseen domains. 
Formally, given a set of labeled source domains $\mathcal{D}_s = \{(x_i, y_i)\}_{i=1}^{N_s}$, where $x_i$ represents the input image and $y_i$ represents the corresponding pixel-wise label, the goal is to train a model $f_{\boldsymbol{\theta}}$ parameterized by $\boldsymbol{\theta}$ that performs well on unseen target domains $\mathcal{D}_t = \{x_j\}_{j=1}^{N_t}$, where the labels for $\mathcal{D}_t$ are not available during training.
The optimization objective for DGSS can be written as:

\begin{equation}
\min_{\boldsymbol{\theta}} \mathbb{E}_{(x_i, y_i) \sim \mathcal{D}_s} \left[ \mathcal{L}(f_{\boldsymbol{\theta}}(x_i), y_i) \right], \label{ERM_S}
\end{equation}
where $\mathcal{L}$ is the segmentation loss (e.g., cross-entropy) that evaluates the difference between the predicted segmentation map $f_{\boldsymbol{\theta}}(x_i)$ and the ground truth $y_i$. The challenge lies in ensuring that the learned model $f_{\boldsymbol{\theta}}$ generalizes well to unseen target domains $\mathcal{D}_t$, which can be written as a generalization objective:

\begin{equation}
\min_{\boldsymbol{\theta}} \mathbb{E}_{x_j \sim \mathcal{D}_t} \left[ \mathcal{L}(f_{\boldsymbol{\theta}}(x_j), y_j^*) \right],
\end{equation}

where $y_j^*$ represents the true (but unknown) labels for the target domain $\mathcal{D}_t$. Since the labels $y_j^*$ are not available, the optimization focuses on learning domain-invariant representations in $\boldsymbol{\theta}$, enabling strong performance across both seen and unseen domains.

\noindent \textbf{Vision Foundation Models} (VFMs), such as CLIP\cite{clip}, MAE\cite{he2022mae}, SAM\cite{kirillov2023sam}, EVA02\cite{fang2023eva02}, and DINOv2\cite{oquab2023dinov2}, almostly use the Vision Transformer (ViT) architecture. 
ViT typically consist of $L$ stacked blocks, each containing two main submodules: multi-head attention (MHA) and a feed-forward network (FFN). 
Specifically, the attention score for each head is calculated as: 
$
\text{MHA}(X) = \text{Concat}(\text{head}_1, \dots, \text{head}_h)\theta_o,
$
$
\quad \text{head}_i = \text{Softmax}\left(\frac{X\theta_{q_i} (X\theta_{k_i})^T}{\sqrt{d_h}}\right)X\theta_{v_i},
$
where $\theta_o$ is the output projection matrix, and $\theta_{q_i}$, $\theta_{k_i}$, and $\theta_{v_i}$ represent the query, key, and value projections for head $i$. The FFN consists of two linear layers with a ReLU activation:
$
\text{FFN}(X) = \text{ReLU}(X\theta_{ffn_1} + \theta_{b_1})\theta_{ffn_2} + \theta_{b_2}.
$
Both the MHA and FFN are followed by residual connections and layer normalization. 
Let $\theta$ denote the set of the parameters of those VFMs that we aim to fine-tune:
\begin{equation}
\boldsymbol{\theta} = [\theta_Q^{(1)}, \theta_K^{(1)}, \theta_V^{(1)}, \theta_{\text{FFN}}^{(1)}, \dots, \theta_Q^{(L)}, \theta_K^{(L)}, \theta_V^{(L)}, \theta_{\text{FFN}}^{(L)}]^\top,
\end{equation}
where $\theta_Q^{(l)} = [\theta_{q_i}, ..., \theta_{q_h}]$
, and so are $\theta_K^{(l)}$ and $\theta_K^{(l)}$.
\noindent \textbf{Fisher Information Matrix (FIM)}  is a fundamental concept in statistical estimation theory\cite{fisher1922mathematical}, which measures the amount of information that an observable random variable carries about an unknown parameter\cite{achille2019information}. In the context of neural networks, the FIM provides insights into the sensitivity of the loss function with respect to the model parameters\cite{matena2022merging}, reflecting the curvature of the loss landscape\cite{rame2022fishr}. Mathematically, the FIM $\mathbf{F}_{\boldsymbol{\theta}}$ is defined as:
\begin{equation}
\mathbf{F}_{\boldsymbol{\theta}} = \mathbb{E}_{x} \left[ \mathbb{E}_{y \sim f_\theta(y|x)} \  \nabla_{\boldsymbol{\theta}} \mathcal{L}(f_{\boldsymbol{\theta}}(x), y) \cdot \nabla_{\boldsymbol{\theta}} \mathcal{L}(f_{\boldsymbol{\theta}}(x), y)^\top \right], \label{eq_fim}
\end{equation}
where $\mathbf{F}_{\boldsymbol{\theta}} \in \mathbb{R}^{|{\boldsymbol{\theta}}| \times |{\boldsymbol{\theta}}|}$ is the symmetrical matrix, $\nabla_{\boldsymbol{\theta}} \mathcal{L}(\cdot)$ denotes the gradient of the loss function to the parameters. 
Intuitively, the Fisher Information Matrix captures how much changing the parameters affects the model's output, thus quantifying the ``informativeness" of the parameters.

\subsection{Motivation}
\begin{figure}[t]
    \begin{center}
    \centering 
    \includegraphics[width=0.45\textwidth]{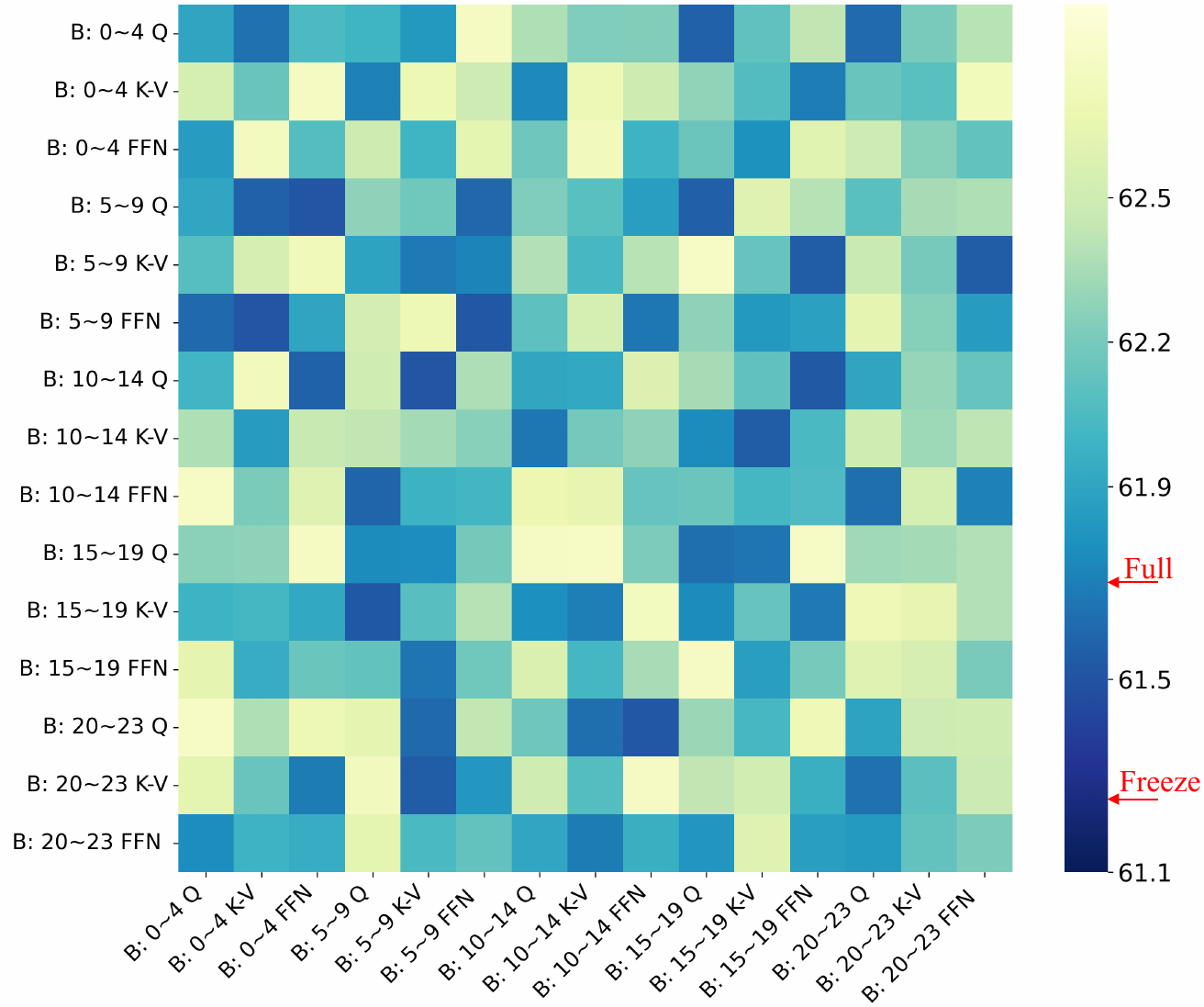} 
    \end{center} 
    \setlength{\abovecaptionskip}{-0.3cm}
    \caption{Observations of fine-tuning different VFM layers for DGSS experiments using DINOV2-large under GTA → Cityscapes + BDD100K + Mapillary. It shows that fine-tuning different layers has different effects on the generalization performance of the VFMs. B means blocks.} 
    \label{Motivation}
\vspace{-0.5cm}
\end{figure}

This study is motivated by an intriguing experimental observation. 
We group the $\theta_Q$, $\theta_K$, $\theta_V$, and $\theta_{\text{FFN}}$ components from different blocks of DINOv2 separately and fine-tune all possible pairwise combinations of these groups, and analyzed their impact on model generalization, as shown in Fig.~\ref{Motivation}. 
We found that tuning specific layers led to different levels of generalization in VFMs, with some configurations even outperforming a fully tuned model. This suggests that certain parameters are critical for maintaining generalization, while others are key to adapting to new domains and tasks. Based on this, we hypothesize that VFMs contain domain-sensitive parameters suited for specific tasks and domains, while other parameters remain broadly generalizable.
Consequently, we propose identifying and fine-tuning these domain-sensitive parameters, enhancing adaptability in DGSS for improved cross-domain performance.
 
\subsection{FisherTune}
In this section, we present our FisherTune, fine-tuning ViT-based Vision Foundation Models (VFMs) guided by the Fisher Information Matrix (FIM) while preserving their generalization strengths.
Our idea is to use FIM to find task- and domain-sensitive parameters in $\boldsymbol{\theta}$ and fine-tune these sensitive parameters to improve the generalization ability of the model on unseen domains while maintaining the pre-trained knowledge of VFMs.

\subsubsection{Domain-Related FIM}

\noindent \textbf{Domain-Related FIM}.
In Eq.~\ref{eq_fim}, FIM quantifies the importance of model parameters for the current task by measuring the sensitivity of parameters to model output. 
However, FIM can not sufficiently capture the behavior of parameters in cross-domain scenes, especially in DGSS tasks, where the sensitivity of different parameters to varying data distributions may differ significantly. 
For the DGSS task, we need a parameters estimation metric that can capture the variation of parameters across different domains.

To address this issue, we propose to calculate the Fisher information change $\Delta \mathbf{F}_{\boldsymbol{\theta}}$ between different data domains (seen domain and simulated unseen domain) to measure the sensitivity difference of parameters across domains.
Formally, given the seen single-source domain $\mathcal{D}_s = \{(x, y)\}$, the $\Delta \mathbf{F}_{\boldsymbol{\theta}}$ is calculated as:
\begin{equation}
\Delta \mathbf{F}_{\boldsymbol{\theta}}  = \frac{|\mathbf{F}_{\boldsymbol{\theta}}(x,y) - \mathbf{F}_{\boldsymbol{\theta}}(x',y)|}  {\min(\mathbf{F}_{{\theta_{i}}}(x), \mathbf{F}_{{\theta_{i}}}(x')) + \epsilon}, \label{delta_F}
\end{equation}
where \( \mathbf{F}_{\boldsymbol{\theta}}(x,y) \) and \( \mathbf{F}_{\boldsymbol{\theta}}(x',y) \)  is the FIM for in the seen and simulated unseen domain. $\epsilon = 1 \times 10^{-8}$ is a small constant to prevent division by zero.
The numerator, \(\left| \mathbf{F}_{\boldsymbol{\theta}}(x, y) - \mathbf{F}_{\boldsymbol{\theta}}(x', y) \right|\), computes the difference between the FIM across domains, reflecting the model's varying sensitivity to parameter changes in different environments or data distributions.
The denominator, \(\min(\mathbf{F}_{{\theta_{i}}}(x), \mathbf{F}_{{\theta_{i}}}(x'))\), normalizes this difference to ensure the relative nature of the metric.
A higher $\Delta \mathbf{F}_{i}$ indicates that the parameter is more sensitive to domain changes.

\begin{figure}[t]
    \begin{center}
    \centering 
    \includegraphics[width=0.5\textwidth]{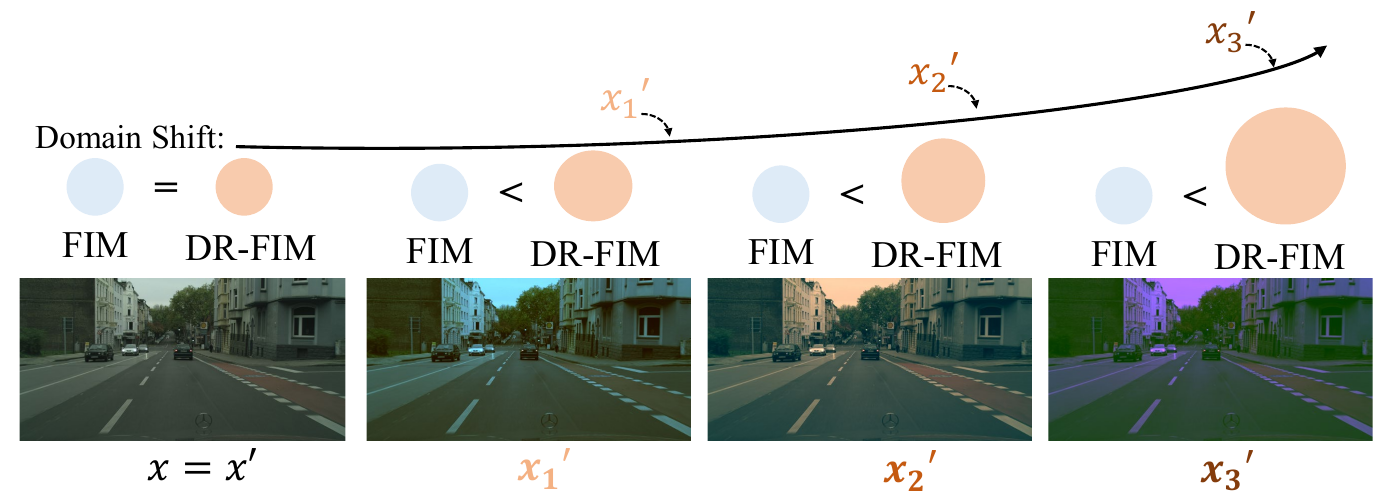} 
    \end{center} 
    \setlength{\abovecaptionskip}{-0.3cm}
    \caption{Comparison of FIM and DR-FIM under different degrees of domain shift. The size of the circle indicates the value. It shows that DR-FIM is a generalization of FIM as it additionally considers the cross-domain sensitivity of parameters.} 
    \label{DR-FIM}
\vspace{-0.5cm}
\end{figure}

To simulate an unseen domain sample \( x' \) from a seen domain  \( x \),  we leverage the uncertainty modeling method inspired by \cite{li2022uncertainty}.
Specifically, the unseen domain feature \( x' \) is simulated by modifying the feature statistics (mean and variance) of the seen domain sample \( x \). The perturbed mean is generated as,
$
\alpha(x) = \mu(x) + \epsilon_\mu \Sigma_\mu(x)$, where \( \mu(x) \) represents the mean of the feature, \( \Sigma_\mu(x) \) is the uncertainty estimation for the mean, and \( \epsilon_\mu \sim \mathcal{N}(0, 1) \) is noise sampled from a standard normal distribution.
Next, the perturbed variance is generated as,
$
\beta(x) = \sigma(x) + \epsilon_\sigma \Sigma_\sigma(x)
$, where \( \sigma(x) \) is the standard deviation of the feature, \( \Sigma_\sigma(x) \) is the uncertainty estimation for the standard deviation, and \( \epsilon_\sigma \sim \mathcal{N}(0, 1) \).
Using the perturbed mean \( \alpha(x) \) and variance \( \beta(x) \), the unseen domain sample \( x' \) is generated with the following formula,
\begin{equation}
x' = \beta(x) \cdot \frac{x - \mu(x)}{\sigma(x)} + \alpha(x). \label{unseen_x}
\end{equation}

Using \(\Delta \mathbf{F}_{\boldsymbol{\theta}}\), we introduce a unified metric, Domain-Related FIM (DR-FIM), to account for both task-sensitive and domain-sensitive parameters as, 
\begin{equation}
\mathbf{DRF}_{\theta} = \underbrace{\mathbf{F}_{\theta}(x, y)}_{\text{task-sensitive}} + \underbrace{ e^{-(\epsilon_\mu + \epsilon_\sigma)} \frac{|\mathbf{F}_{\boldsymbol{\theta}}(x,y) - \mathbf{F}_{\boldsymbol{\theta}}(x',y)|} {\min(\mathbf{F}_{{\theta_{i}}}(x), \mathbf{F}_{{\theta_{i}}}(x')) + \epsilon}}_{\text{domain-sensitive}}. \label{DRF}
\end{equation}
The $\text{DRF}_{\theta}$ is a linear combination of \(\mathbf{F}_{\boldsymbol{\theta}}\) and \(\Delta \mathbf{F}_{\boldsymbol{\theta}}\), 
and the combination coefficients are determined by domain shift control factors \(\epsilon_\mu\) and \(\epsilon_\sigma\). 
When \(\epsilon_\mu\) and \(\epsilon_\sigma\) are large, the simulated domain shift is significant, and \(\Delta \mathbf{F}_{\boldsymbol{\theta}}\) is scaled appropriately to balance with \( \mathbf{F}_{\boldsymbol{\theta}}\). 
The relationship between the simulated domain shift and the numerical values of DR-FIM and FIM is shown in Fig.~\ref{DR-FIM}.

\subsubsection{Stable Estimation of DR-FIM}

Although $ \mathbf{DRF}_{\boldsymbol{\theta}}$ provides an estimation of the domain sensitivity of VFMs parameters, the dimensions of the parameters list ${\boldsymbol{\theta}}$ are very high, which makes it impractical to directly calculate in computation and storage, \textit{i.e.}, \( \boldsymbol O(|{\boldsymbol{\theta}}|^2) \). 
Therefore, it is necessary to approximate the FIM to reduce the computational complexity.

\noindent \textbf{Diagonal Approximation}. Following \cite{matena2022merging}, by assuming that the off-diagonal elements are negligible, 
the FIM can be efficiently approximated by using a diagonal approximation, \textit{i.e.},
\begin{equation}
\hat{ \mathbf F}_{\theta} = \text{diag}({ \mathbf F}_{\theta_1}, {\mathbf F}_{\theta_2},...,{\mathbf F}_{\theta_{|{\boldsymbol{\theta}}|}}),
\end{equation}
where each individual ${\mathbf F}_{\theta_n}$ can be calculated as,  
\begin{equation}
{\mathbf F}_{\theta_n} = \frac{1}{N} \sum_{i=1}^{N} \mathbb{E}_{y_i \sim f_\theta(y_i|x_i)} \left( \nabla_{{\theta_n}} \mathcal{L}(f_{{\theta}}(x), y_i) \right)^2, \label{FIM_Approx}
\end{equation}
In the above diagonal approximation, only the individual contribution of each parameter to the loss is considered, while the interaction terms between different parameters are ignored. This diagonal approximation effectively simplifies a \( \boldsymbol O(|{\boldsymbol{\theta}}|^2) \) matrix to a vector of length \( \boldsymbol O(|{\boldsymbol{\theta}}|) \), greatly reducing the computational complexity.

\noindent \textbf{Variational Estimation of DR-FIM}. Diagonalization provides an efficient approximation method for computing the FIM. However, due to the differences between the pretraining tasks of the VFM and the DGSS task, the estimated FIM parameters often exhibit high sensitivity, leading to inaccuracies (See Fig.~\ref{qkv_show}). 
To address this issue, we introduce a variational inference approach \cite{blei2017variational, pu2020dual}, treating the fine-tuning model’s parameters \( \boldsymbol{\theta} \) as random variables following a Gaussian distribution. This introduces an additional regularization term into the FIM estimation, helping to learn a smoother prior distribution. Consequently, variational inference stabilizes the gradient update process during FIM calculation, mitigating the instability caused by high gradient noise.

Specifically, assuming that the posterior distribution of the model parameters follows a Gaussian distribution: $q(\boldsymbol{\theta}) = \mathcal{N}(\hat{\boldsymbol{\theta}}, \Lambda^{-1})$, where \( \hat{\boldsymbol{\theta}} \) is the mean of the current parameters estimation, \( \Lambda^{-1} \) is the covariance matrix of the parameters.
To preserve the pre-trained knowledge of VFMs, we introduce the prior parameter distribution as a regularizer to prevent degradation during prediction:
\begin{equation}
p(\boldsymbol{\theta}) = \mathcal{N}(\boldsymbol{\theta}_{\text{pt}}, \tau^2 I), \label{prior}
\end{equation}
where \( {\boldsymbol{\theta}}_{\text{pt}} \) is the pre-trained parameters of VFMs, \( \tau^2 \) is the variance controlling the flexibility of fine-tune parameters, and \( I \) is the identity matrix.
We utilize the variational free energy (also called the evidence lower bound, ELBO \cite{hoffman2013stochastic}) as the loss function for optimizing $\Lambda$,
\begin{equation}
L(\hat{\boldsymbol{\theta}}, \Lambda^{-1}) = \mathbb{E}_{\boldsymbol{\theta} \sim q(\boldsymbol{\theta})} \left[ \mathcal{L}(\boldsymbol{\theta}) \right] + \gamma \, KL(q(\boldsymbol{\theta}) \| p(\boldsymbol{\theta})) ,\label{ELBO_C}
\end{equation}
where \( \gamma \) is the regularization coefficient, controlling the influence of the prior, and \( KL(q(\boldsymbol{\theta}) \| p(\boldsymbol{\theta})) \) is the Kullback-Leibler divergence between the posterior \( q(\boldsymbol{\theta}) \) and the prior \( p(\boldsymbol{\theta}) \).

\noindent \textbf{Connection with DR-FIM}.
To simplify the first term in Eq.~(\ref{ELBO_C}), we perform a second-order Taylor expansion of the loss function \( \mathcal{L}(\boldsymbol{\theta}) \) around the current parameters estimate \( \boldsymbol{\theta} = \hat{\boldsymbol{\theta}} \).
Taking the expectation over the weight distribution \( q(\boldsymbol{\theta}) \),
\(
\mathbb{E}_{\boldsymbol{\theta} \sim q(\boldsymbol{\theta})} \left[ \mathcal{L}(\boldsymbol{\theta}) \right] \approx \mathcal{L}(\hat{\boldsymbol{\theta}}) + \frac{1}{2} \operatorname{Tr} \left( \nabla_{\boldsymbol{\theta}}^2 \mathcal{L}(\hat{\boldsymbol{\theta}}) \Lambda^{-1} \right) .
\)
\noindent
According to the definition of FIM and its connection with the Hessian matrix\cite{fisher1922mathematical}, the FIM can be approximated by the Hessian matrix near \( \hat{\boldsymbol{\theta}} \),
\( \nabla_{\boldsymbol{\theta}}^2 \mathcal{L}(\hat{\boldsymbol{\theta}}) \approx \mathbf{F}_{\boldsymbol{\theta}}. \) 
Thus,
\begin{equation}
\mathbb{E}_{\boldsymbol{\theta} \sim q(\boldsymbol{\theta})} \left[ \mathcal{L}(\boldsymbol{\theta}) \right] \approx \mathcal{L}(\hat{\boldsymbol{\theta}}) + \frac{1}{2} \operatorname{Tr} \left( \mathbf{F}_{\boldsymbol{\theta}} \Lambda^{-1}. \right) \label{elbo_t1}
\end{equation}

The second term in Eq.~(\ref{ELBO_C}), KL divergence between two Gaussian distributions is simplified by,
\begin{equation}
\begin{aligned}
KL(q(\boldsymbol{\theta}) \| p(\boldsymbol{\theta})) = \frac{1}{2} \Big( & \tau^{-2} \operatorname{Tr}( \Lambda^{-1} ) 
+ \tau^{-2} \| \hat{\boldsymbol{\theta}} - {\boldsymbol{\theta}}_{\text{pt}} \|^2 \\
& - k + k \ln \tau^2 + \ln \det \Lambda \Big).
\end{aligned} \label{elbo_t2}
\end{equation}
Substituting the Eq.~(\ref{elbo_t1}) and Eq.~(\ref{elbo_t2}) back into Eq.~(\ref{ELBO_C}), and taking the derivative of the loss function with respect to \( \Lambda \) and we obtain (\textcolor[rgb]{0,0.69,0.94}{\textbf{See Appendix A for detailed derivation}}),
\begin{equation}
\mathbf{F}_{\boldsymbol{\theta}} = \gamma \Lambda - \gamma \tau^{-2} I.
\end{equation}
Then, the DR-FIM defined in Eq.~(\ref{DRF}) is updated as, 
\begin{equation}
\mathbf{DRF}_{\theta} = \gamma \left( \Lambda_x - \tau^{-2} I + e^{-(\epsilon_\mu + \epsilon_\sigma)} \frac{|\Lambda_x - \Lambda_{x'}|}{\min(\Lambda_x, \Lambda_{x'}) + \frac{\epsilon}{\gamma}} \right). \label{delta_f}
\end{equation}
It shows that the DR-FIM can be estimated from the covariance matrix \( \Lambda \), with \( \gamma \) and \( \tau \) as hyperparameters. 
Using Eq.~(\ref{delta_f}) to estimate the DR-FIM has several advantages over using Eq.~(\ref{DRF}) and Eq.~(\ref{FIM_Approx}). 
1) Stability in Estimation: Eq.~(\ref{delta_f}) introduces a more stable estimation mechanism by incorporating prior knowledge from the pre-trained VFMs $p(\boldsymbol{\theta})$ in Eq.~(\ref{prior}) and the posterior distribution $q(\boldsymbol{\theta})$. This approach helps prevent the degradation of FIM estimation caused by the task shift between VFM pre-training tasks and DGSS tasks, ensuring more robust performance across unseen domains.
2) Computational Efficiency: The covariance matrix  \( \Lambda \) can be efficiently computed by directly minimizing the loss $L(\hat{\boldsymbol{\theta}}, \Lambda^{-1})$ in Eq.~(\ref{ELBO_C}) using reparameterization trick\cite{kingma2015variational} and stochastic gradient variational Bayes \cite{achille2019task2vec}, reducing both computational and memory overhead compared to traditional FIM estimation in Eq.~(\ref{FIM_Approx}).



\subsubsection{Training Scheduler of FisherTune}
We follow Rein\cite{Wei_2024_CVPR} which adds a mask decoder to the backbone network of VFMs as a segmentation model for DGSS. 
Different from Rein, we do not modify the backbone structure or add additional adapters.
During training, we first fix the backbone network of VFMs and use the original data to warm-up the decoder to adapt the whole segmentation model to the DGSS task.
After that, we tune the VFMs and decoder by our FisherTune as follows.

In FisherTune, the selection of parameters for fine-tuning is guided by the DR-FIM (\(\mathbf{DRF}_{\boldsymbol{\theta}}\)), which quantifies the sensitivity of parameters to task and domain shifts. To optimize the fine-tuning process, we propose a dynamic training schedule that adjusts the number of trainable parameters based on their DR-FIM values.
At the beginning of training, we fine-tune only the most sensitive \( \delta_{\min}\)\% of parameters, as ranked by \(\mathbf{DRF}_{\boldsymbol{\theta_i}}\). As training progresses, we gradually increase the percentage of fine-tuned parameters, reaching \(\delta_{\max}\)\% by the end. This ensures that the model starts with a focused fine-tuning process, targeting only the most critical parameters, and progressively expands the fine-tuning scope as the model becomes more stable.
Formally, at each training step \(t\), the dynamic threshold \(\mathbf{DRF}_{\text{thresh}}(t)\) is updated as follows:
\begin{equation}
\mathbf{DRF}_{\text{thresh}}(t) = \delta_{min} + (\delta_{\max} - \delta_{\min}) \cdot \exp\left(- \frac{t}{T} \right), \label{choose_Scheduler}
\end{equation}
where \(T\) is the total number of training steps.
Parameters with \(\mathbf{DRF}_{\boldsymbol{\theta}}\) values higher than the threshold \(\mathbf{DRF}_{\text{thresh}}(t)\) will be selected for training.
The detailed fine-tuning process of our FisherTune is in Algorithm~\ref{Algorithm_FisherTune}.


\begin{algorithm}[t]
\caption{FisherTune Process}
\begin{algorithmic}[1]
    \State \textbf{Input:} source dataset $\mathcal{D} = \{(x_i, y_i)\}_{i=1}^{N}$;
    Hyperparameters: regularization coefficient $\gamma$, variance coefficient \( \tau \), warm-up iterations $T_1$, FIM estimation iterations $T_2$, number of tune iterations $T_3$; 
    pretrained VFM $\boldsymbol \theta_{\text{VFM}}$;
    segmentation decoder $\boldsymbol \theta_{\text{dec}}$.
    
    \State \textbf{Step 1: Warm-up decoder:}
    \State Train the decoder $\boldsymbol \theta_{\text{dec}}$ on $\mathcal{D}$ for $T_1$ steps, froze $\boldsymbol \theta_{\text{VFM}}$.
    
    \State \textbf{Step 2: Sampling and DR-FIM Calculation:}
    \For{$t = 1$ to $T_2$}
        \State Sample batch $(x, y) \sim \mathcal{D}$
            
            \State Simulate unseen domain data $x'$ via Eq.~\ref{unseen_x}.
            
            \State Optimize covariance matrix $\Lambda$ via Eq.~\ref{ELBO_C}
            
            \State Estimate DR-FIM using the optimized $\Lambda$ via Eq.~\ref{delta_f}
    \EndFor    
    \State \textbf{Step 3: Parameter Fine-Tuning:}
        \For{$t = 1$ to $T_3$}
            \State Sample a batch $(x, y) \sim \mathcal{D}$
            
            \State Select parameters $\boldsymbol {\hat \theta_{\text{VFM}}} $ via Eq.~\ref{choose_Scheduler}
            
            \State Update the selected $\boldsymbol {\hat \theta_{\text{VFM}}} $  and $\boldsymbol \theta_{\text{dec}}$ via Eq.~\ref{ERM_S}.
        \EndFor
    
    \State \textbf{Output:} Fine-tuned $\boldsymbol \theta_{\text{VFM}}$ and $\boldsymbol \theta_{\text{dec}}$.

\end{algorithmic} \label{Algorithm_FisherTune}
\end{algorithm}


\section{Experiments}
\subsection{Datasets \& Setup \textcolor[rgb]{0,0.69,0.94}{\textbf{See Appendix B.}}} 

\begin{table}[tbp]
  \centering
   \renewcommand{\arraystretch}{1.05}
  \resizebox{\linewidth}{!}{
\begin{tabular}{ccccccc}
\toprule
\multicolumn{7}{c}{GTAV → Cityscapes (Citys) + BDD100K (BDD) + Mapillary (Map)} \\
\midrule
      VFM type & Fine-tune Method & Trainable Params & Citys & BDD   & Map   & Avg. \\
\midrule
\multirow{5}[1]{*}{\makecell{CLIP \cite{clip} \\ (ViT-Large)}} & Full  & 304.20M & 51.3  & 47.6  & 54.3  & 51.1  \\
      & Freeze & 0M    & 53.7  & 48.7  & 55.0  & 52.5  \\
      & LoRA \cite{hu2021lora} & 0.79M & 54.0 & 49.8 & 55.1 & 53.0  \\
      & VPT \cite{jia2022visual}   & 3.69M & 54.0 & 51.8 & 57.5 & 54.4 \\
      & Rein \cite{Wei_2024_CVPR} & 2.99M & 57.1  & \cellcolor[rgb]{ .886,  .937,  .855}54.7  & \cellcolor[rgb]{ .886,  .937,  .855}60.5  & \cellcolor[rgb]{ .886,  .937,  .855}57.4  \\
      & VQT \cite{tu2023visual} & 3.01M & 54.3 & 51.2 & 56.7 & 55.3  \\
      & ChildTune \cite{xu2021raise} & 15.21M & \cellcolor[rgb]{ .886,  .937,  .855}57.9 & 53.4 & 58.2 & 56.5  \\
      & Ours & 15.21M & \cellcolor[rgb]{ .741,  .843,  .933} \textbf{59.2} & \cellcolor[rgb]{ .741,  .843,  .933} \textbf{57.5} & \cellcolor[rgb]{ .741,  .843,  .933} \textbf{61.0} & \cellcolor[rgb]{ .741,  .843,  .933} \textbf{59.2} \\
\midrule
\multirow{5}[1]{*}{\makecell{MAE \cite{he2022mae} \\ (Huge))}} & Full  & 304.20M & 53.7  & \cellcolor[rgb]{ .886,  .937,  .855}50.8  & 58.1  & 54.2  \\
      & Freeze & 0M  & 43.3 & 37.8 & 48.0 & 43.0  \\
      & LoRA \cite{hu2021lora} & 0.79M & 44.6 & 38.4 & 52.5 & 45.2  \\
      & VPT \cite{jia2022visual}   & 3.69M  & 52.7 & 50.2 & 57.6 & 53.5 \\
      & Rein \cite{Wei_2024_CVPR}  & 2.99M & 55.0  & 49.3  & \cellcolor[rgb]{ .886,  .937,  .855}58.6  & 54.3  \\
      & VQT \cite{tu2023visual} & 3.01M & 53.3 & 50.3 & 57.7 & 53.8  \\
      & ChildTune \cite{xu2021raise} & 15.21M & \cellcolor[rgb]{ .886,  .937,  .855}55.4 & 50.6 & 58.1 & \cellcolor[rgb]{ .886,  .937,  .855}54.7  \\
      & Ours & 15.21M & \cellcolor[rgb]{ .741,  .843,  .933} \textbf{56.6} & \cellcolor[rgb]{ .741,  .843,  .933} \textbf{51.9} & \cellcolor[rgb]{ .741,  .843,  .933} \textbf{59.7}  & \cellcolor[rgb]{ .741,  .843,  .933} \textbf{56.1} \\
\midrule
\multirow{5}[2]{*}{\makecell{SAM \cite{kirillov2023sam} \\ (Huge)}} & Full  & 632.18M & 57.6  & 51.7  & 61.5  & 56.9  \\
      & Freeze & 0M    & 57.0  & 47.1  & 58.4  & 54.2  \\
      & LoRA \cite{hu2021lora} & 0.79M & 57.4 & 47.7 & 58.4 & 54.5  \\
      & VPT \cite{jia2022visual}   & 3.69M & 56.3  & 52.7  & 57.8  & 55.6  \\
      & Rein \cite{Wei_2024_CVPR}  & 2.99M & 59.6  & 52.0  & \cellcolor[rgb]{ .886,  .937,  .855}62.1  & \cellcolor[rgb]{ .886,  .937,  .855}57.9  \\
      & VQT \cite{tu2023visual} & 3.01M & 56.7 & \cellcolor[rgb]{ .886,  .937,  .855}53.9 & 59.3 & 56.6  \\
      & ChildTune \cite{xu2021raise} & 15.21M & \cellcolor[rgb]{ .886,  .937,  .855}60.8 & 49.6 & 61.2 & 57.2   \\
      & Ours & 15.21M & \cellcolor[rgb]{ .741,  .843,  .933} \textbf{60.9} & \cellcolor[rgb]{ .741,  .843,  .933} \textbf{54.4} & \cellcolor[rgb]{ .741,  .843,  .933} \textbf{63.9}  & \cellcolor[rgb]{ .741,  .843,  .933} \textbf{59.7} \\
\midrule
\multirow{7}[2]{*}{\makecell{EVA02 \cite{eva02} \\
(Large)}} & Full  & 304.20M & 62.1  & 56.2  & \cellcolor[rgb]{ .886,  .937,  .855}64.6  & 60.9  \\
      & LoRA \cite{hu2021lora}  & 0.79M & 55.5  & 52.7  & 58.3  & 55.5  \\
      & AdaptFormer\cite{chen2022adaptformer} & 3.17M & 63.7  & 59.9  & 64.2  & 62.6  \\
      & VPT\cite{jia2022visual}   & 3.69M & 62.2  & 57.7  & 62.5  & 60.8  \\
      & Rein\cite{Wei_2024_CVPR}  & 2.99M & \cellcolor[rgb]{ .886,  .937,  .855}65.3  & \cellcolor[rgb]{ .886,  .937,  .855}61.1  & 63.9  & \cellcolor[rgb]{ .886,  .937,  .855}63.4  \\
      & VQT\cite{tu2023visual} & 3.01M & 61.3  & 55.1  & 62.2  & 59.5  \\
      & ChildTune\cite{xu2021raise} & 15.21M & 61.6  & 59.3  & 62.3  & 61.1  \\
      & Ours & 15.21M & \cellcolor[rgb]{ .741,  .843,  .933} \textbf{65.8} & \cellcolor[rgb]{ .741,  .843,  .933} \textbf{61.5} & \cellcolor[rgb]{ .741,  .843,  .933} \textbf{66.0} & \cellcolor[rgb]{ .741,  .843,  .933} \textbf{64.4} \\
\midrule
\multirow{7}[2]{*}{\makecell{DINOv2\cite{oquab2023dinov2} \\
(ViT-Large)}} & Full  & 304.20M & 63.7  & 57.4  & 64.2  & 61.7  \\
      & LoRA\cite{hu2021lora}  & 0.79M  & 65.2  & 58.3  & 64.6  & 62.7  \\
      & AdaptFormer\cite{chen2022adaptformer} & 3.17M & 64.9  & 59.0  & 64.2  & 62.7  \\
      & VQT \cite{jia2022visual}  & 3.01M & 64.6  & 59.0  & 65.7  & 63.1  \\
      & Rein\cite{Wei_2024_CVPR}  & 2.99M  & \cellcolor[rgb]{ .886,  .937,  .855}66.4  & \cellcolor[rgb]{ .886,  .937,  .855}60.4  & \cellcolor[rgb]{ .886,  .937,  .855}66.1  & \cellcolor[rgb]{ .886,  .937,  .855}64.3  \\
      & ChildTune\cite{xu2021raise} & 15.21M & 65.6  & 59.3  & 65.3  & 63.4  \\
      & Ours  & 15.21M  & \cellcolor[rgb]{ .741,  .843,  .933} \textbf{68.2} & \cellcolor[rgb]{ .741,  .843,  .933} \textbf{63.3} & \cellcolor[rgb]{ .741,  .843,  .933} \textbf{68.0} & \cellcolor[rgb]{ .741,  .843,  .933} \textbf{66.5} \\
\midrule
EVA02  & VLTSeg~\cite{hummer2024strong} & 304.2M      & 65.3  & 58.3  & 66.0  & 63.2  \\
DINOV2 & SDT~\cite{yi2024learning} &  6.94M  & 68.1  & \cellcolor[rgb]{ .886,  .937,  .855} 61.6  & 67.7  & 65.8  \\
CLIP+SAM & CLOUDS ~\cite{pak2024textual} &    304.2M   & 60.2  & 57.4  & 67.0  & 61.5  \\
EVA02 & tqdm ~\cite{pak2024textual} &   304.2M    & \cellcolor[rgb]{ .741,  .843,  .933} \textbf{68.9} & 59.2  & \cellcolor[rgb]{ .741,  .843,  .933} \textbf{70.1} & \cellcolor[rgb]{ .886,  .937,  .855} 66.1  \\
EVA02 & Ours  &   15.21M    & 65.8  & 61.5  & 66.0  & 64.4  \\
DINOV2 & Ours  &  15.21M     & \cellcolor[rgb]{ .886,  .937,  .855}68.2  & \cellcolor[rgb]{ .741,  .843,  .933}\textbf{63.3} & \cellcolor[rgb]{ .886,  .937,  .855}68.7  & \cellcolor[rgb]{ .741,  .843,  .933}\textbf{66.6} \\
\bottomrule
\end{tabular}
}
 \setlength{\abovecaptionskip}{0.05 cm}
  \caption{Performance and Trainable Parameters 
Comparison with the proposed FisherTune across Multiple VFMs as Backbones under the GTAV → Cityscapes (Citys) + BDD100K (BDD) + Mapillary (Map) generalization setting.}
  \label{exp_vfm_all}
  \vspace{-0.6cm}
\end{table}

\begin{table*}[t]
  \centering
   \renewcommand{\arraystretch}{1.05}
  \resizebox{\linewidth}{!}{
\begin{tabular}{ccccccccccccccccccccccc}
\toprule
      & \multicolumn{22}{c}{Cityscapes → BDD100K} \\
\midrule
      & Fine-tune Method & Trainable Params & road  & side. & build. & wall  & fence & pole  & light & sign  & vege  & terr. & sky   & pers. & rider & car   & truck & bus   & train & moto. & bicy. & mIoU \\
\midrule
\multirow{6}[2]{*}{DINOv2} & Full  & 304.20M & 89.0  & 44.5  & 89.6  & 51.1  & 46.4  & 49.2  & \cellcolor[rgb]{ .741,  .843,  .933} \textbf{60.0} & 38.9  & 89.1  & 47.5  & \cellcolor[rgb]{ .886,  .937,  .855} 91.7  & \cellcolor[rgb]{ .741,  .843,  .933} \textbf{75.8} & \cellcolor[rgb]{ .886,  .937,  .855} 48.2  & 91.7  & 52.5  & \cellcolor[rgb]{ .886,  .937,  .855} 82.9  & \cellcolor[rgb]{ .741,  .843,  .933} \textbf{81.0} & 30.4  & 49.9  & 63.7  \\
      & Freeze & 0M    & 92.1  & 55.2  & 90.2  & 57.2  & 48.5  & 49.5  & 56.7  & \cellcolor[rgb]{ .886,  .937,  .855} 47.7  & 89.3  & 47.8  & 91.1  & \cellcolor[rgb]{ .886,  .937,  .855} 74.2  & 46.7  & 92.2  & 62.6  & 77.5  & 47.7  & 29.6  & 47.2  & 63.3  \\
      & REIN \cite{Wei_2024_CVPR} & 2.99M & \cellcolor[rgb]{ .741,  .843,  .933} \textbf{92.4} & \cellcolor[rgb]{ .741,  .843,  .933} \textbf{59.1} & \cellcolor[rgb]{ .886,  .937,  .855} 90.7  & 58.3  & \cellcolor[rgb]{ .741,  .843,  .933} \textbf{53.7} & 51.8  & 58.2  & 46.4  & 89.8  & \cellcolor[rgb]{ .886,  .937,  .855} 49.4  & 90.8  & 73.9  & 43.3  & 92.3  & \cellcolor[rgb]{ .886,  .937,  .855} 64.3  & 81.6  & 70.9  & 40.4  & 54.0  & \cellcolor[rgb]{ .886,  .937,  .855} 66.4  \\
      & VQT \cite{tu2023visual} & 3.01M & 88.3  & 49.9  & 85.9  & 50.7  & 47.9  & 44.3  & 55.6  & 39.2  & 86.1  & 42.8  & 87.5  & 71.3  & 45.4  & 89.4  & 53.5  & 82.6  & \cellcolor[rgb]{ .886,  .937,  .855} 74.9  & \cellcolor[rgb]{ .741,  .843,  .933} \textbf{46.1} & \cellcolor[rgb]{ .741,  .843,  .933} \textbf{57.4} & 63.1  \\
      & ChildTune \cite{xu2021raisechildlargelanguage} & 15.21M & 92.1  & \cellcolor[rgb]{ .886,  .937,  .855} 56.1  & \cellcolor[rgb]{ .741,  .843,  .933} \textbf{91.0} & \cellcolor[rgb]{ .886,  .937,  .855} 58.8  & 46.9  & \cellcolor[rgb]{ .886,  .937,  .855} 52.0  & 58.6  & 47.2  & \cellcolor[rgb]{ .886,  .937,  .855} 90.8  & 47.9  & \cellcolor[rgb]{ .741,  .843,  .933} \textbf{93.3} & 72.0  & 47.1  & \cellcolor[rgb]{ .741,  .843,  .933} \textbf{93.0} & 63.9  & 76.2  & 47.9  & 28.8  & 48.3  & 63.8  \\
      & Ours  & 15.21M & \cellcolor[rgb]{ .886,  .937,  .855} 92.1  & 55.4  & 90.2  & \cellcolor[rgb]{ .741,  .843,  .933} \textbf{58.9} & \cellcolor[rgb]{ .886,  .937,  .855} 50.9  & \cellcolor[rgb]{ .741,  .843,  .933} \textbf{54.5} & \cellcolor[rgb]{ .886,  .937,  .855} 59.8  & \cellcolor[rgb]{ .741,  .843,  .933} \textbf{49.1} & \cellcolor[rgb]{ .741,  .843,  .933} \textbf{92.5} & \cellcolor[rgb]{ .741,  .843,  .933} \textbf{52.8} & 91.0  & 73.7  & \cellcolor[rgb]{ .741,  .843,  .933} \textbf{51.5} & \cellcolor[rgb]{ .886,  .937,  .855} 92.7  & \cellcolor[rgb]{ .741,  .843,  .933} \textbf{67.4} & \cellcolor[rgb]{ .741,  .843,  .933} \textbf{82.9} & 72.8  & \cellcolor[rgb]{ .886,  .937,  .855} 44.3  & \cellcolor[rgb]{ .886,  .937,  .855} 54.1  & \cellcolor[rgb]{ .741,  .843,  .933} \textbf{67.7} \\
\midrule
\multirow{6}[2]{*}{EVA02} & Full  & 304.20M & 89.3  & 46.9  & 89.9  & 47.7  & 45.6  & 50.1  & \cellcolor[rgb]{ .886,  .937,  .855} 56.8  & 42.2  & 88.8  & 48.4  & 89.9  & 75.8  & 49.0  & 90.5  & 45.3  & 69.2  & 55.9  & 44.4  & 55.1  & 62.1  \\
      & REIN \cite{Wei_2024_CVPR} & 0M    & \cellcolor[rgb]{ .741,  .843,  .933} \textbf{93.1} & \cellcolor[rgb]{ .741,  .843,  .933} \textbf{52.7} & 88.0  & 47.4  & 31.1  & 41.7  & 46.0  & 39.6  & 85.7  & 41.4  & 89.5  & 67.5  & 39.7  & 89.0  & 47.0  & \cellcolor[rgb]{ .886,  .937,  .855} 72.8  & 46.3  & 19.2  & 35.2  & 56.5  \\
      & VQT \cite{tu2023visual} & 2.99M & 91.7  & \cellcolor[rgb]{ .886,  .937,  .855} 51.8  & 90.1  & \cellcolor[rgb]{ .741,  .843,  .933} \textbf{52.8} & \cellcolor[rgb]{ .886,  .937,  .855} 48.4  & 48.2  & 56.0  & 42.0  & 89.1  & 44.1  & 90.2  & 74.2  & 47.0  & 91.1  & \cellcolor[rgb]{ .741,  .843,  .933} \textbf{54.5} & \cellcolor[rgb]{ .741,  .843,  .933} \textbf{84.1} & \cellcolor[rgb]{ .741,  .843,  .933} \textbf{78.9} & \cellcolor[rgb]{ .741,  .843,  .933} \textbf{47.2} & 59.4  & \cellcolor[rgb]{ .886,  .937,  .855} 65.3  \\
      & ChildTune \cite{xu2021raisechildlargelanguage} & 3.01M & 90.1  & 46.6  & \cellcolor[rgb]{ .886,  .937,  .855} 91.1  & 46.9  & 46.4  & \cellcolor[rgb]{ .886,  .937,  .855} 51.7  & 56.5  & 43.2  & 89.3  & 49.6  & \cellcolor[rgb]{ .886,  .937,  .855} 92.3  & 75.0  & 50.3  & 90.3  & 44.6  & 71.8  & \cellcolor[rgb]{ .886,  .937,  .855} 57.4  & 44.0  & 55.8  & 62.8  \\
      & ChildTune & 15.21M & 91.4  & 50.7  & 88.9  & 47.9  & 47.4  & \cellcolor[rgb]{ .741,  .843,  .933} \textbf{54.6} & 56.3  & \cellcolor[rgb]{ .741,  .843,  .933} \textbf{45.9} & \cellcolor[rgb]{ .886,  .937,  .855} 91.2  & \cellcolor[rgb]{ .886,  .937,  .855} 50.0  & 91.2  & \cellcolor[rgb]{ .886,  .937,  .855} 76.1  & \cellcolor[rgb]{ .741,  .843,  .933} \textbf{52.2} & \cellcolor[rgb]{ .886,  .937,  .855} 92.3  & \cellcolor[rgb]{ .886,  .937,  .855} 48.0  & 69.3  & 55.2  & 43.9  & \cellcolor[rgb]{ .886,  .937,  .855} 59.8  & 63.8  \\
      & Ours  & 15.21M & \cellcolor[rgb]{ .886,  .937,  .855} 92.6  & 49.9  & \cellcolor[rgb]{ .741,  .843,  .933} \textbf{95.9} & \cellcolor[rgb]{ .886,  .937,  .855} 51.1  & \cellcolor[rgb]{ .741,  .843,  .933} \textbf{53.0} & 50.8  & \cellcolor[rgb]{ .741,  .843,  .933} \textbf{59.8} & \cellcolor[rgb]{ .886,  .937,  .855} 45.7  & \cellcolor[rgb]{ .741,  .843,  .933} \textbf{92.9} & \cellcolor[rgb]{ .741,  .843,  .933} \textbf{54.6} & \cellcolor[rgb]{ .741,  .843,  .933} \textbf{94.0} & \cellcolor[rgb]{ .741,  .843,  .933} \textbf{83.5} & \cellcolor[rgb]{ .886,  .937,  .855} 52.2  & \cellcolor[rgb]{ .741,  .843,  .933} \textbf{93.9} & 45.1  & 69.4  & 57.1  & \cellcolor[rgb]{ .886,  .937,  .855} 47.2  & \cellcolor[rgb]{ .741,  .843,  .933} \textbf{62.4} & \cellcolor[rgb]{ .741,  .843,  .933} \textbf{65.8} \\
\midrule
      & \multicolumn{22}{c}{Cityscapes → ACDC} \\
\midrule
\multirow{6}[2]{*}{DINOv2} & Full  & 304.20M & 92.8  & 75.0  & 87.4  & 55.7  & 54.1  & 55.6  & 71.2  & 69.6  & 82.4  & 56.0  & 92.2  & 66.8  & 45.6  & 89.0  & 79.7  & \cellcolor[rgb]{ .886,  .937,  .855} 87.9  & 87.5  & 51.4  & 62.7  & 71.7  \\
      & Freeze & 0M    & 86.0  & 68.1  & 80.2  & 52.4  & 47.8  & 48.2  & 65.5  & 65.3  & 80.0  & 54.7  & 86.2  & 65.0  & 44.9  & 86.4  & 73.3  & 80.5  & 86.9  & 50.1  & 60.9  & 67.5  \\
      & REIN \cite{Wei_2024_CVPR} & 2.99M & \cellcolor[rgb]{ .886,  .937,  .855} 94.6  & \cellcolor[rgb]{ .886,  .937,  .855} 78.3  & \cellcolor[rgb]{ .886,  .937,  .855} 92.0  & \cellcolor[rgb]{ .741,  .843,  .933} \textbf{61.9} & \cellcolor[rgb]{ .886,  .937,  .855} 55.0  & \cellcolor[rgb]{ .886,  .937,  .855} 64.8  & \cellcolor[rgb]{ .886,  .937,  .855} 73.8  & \cellcolor[rgb]{ .886,  .937,  .855} 72.7  & \cellcolor[rgb]{ .741,  .843,  .933} \textbf{88.4} & \cellcolor[rgb]{ .741,  .843,  .933} \textbf{67.4} & \cellcolor[rgb]{ .886,  .937,  .855} 95.4  & \cellcolor[rgb]{ .741,  .843,  .933} \textbf{77.1} & \cellcolor[rgb]{ .741,  .843,  .933} \textbf{60.2} & \cellcolor[rgb]{ .886,  .937,  .855} 92.6  & \cellcolor[rgb]{ .886,  .937,  .855} 84.1  & 86.9  & \cellcolor[rgb]{ .886,  .937,  .855} 92.5  & \cellcolor[rgb]{ .741,  .843,  .933} \textbf{67.6} & \cellcolor[rgb]{ .741,  .843,  .933} \textbf{68.6} & \cellcolor[rgb]{ .741,  .843,  .933} \textbf{77.6} \\
      & VQT \cite{tu2023visual} & 3.01M & 93.3  & 76.4  & 89.2  & 55.0  & 53.9  & 53.9  & 72.0  & 67.3  & 83.4  & 55.3  & 95.1  & 67.7  & 47.0  & 90.5  & 81.6  & 86.3  & 88.2  & 50.1  & 61.9  & 72.0  \\
      & ChildTune \cite{xu2021raisechildlargelanguage} & 15.21M & 92.9  & 72.8  & 84.7  & 56.6  & 54.1  & 56.8  & 70.9  & 67.7  & 82.3  & 55.7  & 93.6  & 65.9  & 45.3  & 89.6  & 77.6  & 87.8  & 87.0  & 52.5  & 62.2  & 71.4  \\
      & Ours  & 15.21M & \cellcolor[rgb]{ .741,  .843,  .933} \textbf{95.6} & \cellcolor[rgb]{ .741,  .843,  .933} \textbf{79.0} & \cellcolor[rgb]{ .741,  .843,  .933} \textbf{96.5} & \cellcolor[rgb]{ .886,  .937,  .855} 60.5  & \cellcolor[rgb]{ .741,  .843,  .933} \textbf{58.3} & \cellcolor[rgb]{ .741,  .843,  .933} \textbf{64.9} & \cellcolor[rgb]{ .741,  .843,  .933} \textbf{75.6} & \cellcolor[rgb]{ .741,  .843,  .933} \textbf{77.7} & \cellcolor[rgb]{ .886,  .937,  .855} 85.0  & \cellcolor[rgb]{ .886,  .937,  .855} 61.3  & \cellcolor[rgb]{ .741,  .843,  .933} \textbf{98.6} & \cellcolor[rgb]{ .886,  .937,  .855} 73.6  & \cellcolor[rgb]{ .886,  .937,  .855} 51.5  & \cellcolor[rgb]{ .741,  .843,  .933} \textbf{94.8} & \cellcolor[rgb]{ .741,  .843,  .933} \textbf{85.4} & \cellcolor[rgb]{ .741,  .843,  .933} \textbf{94.7} & \cellcolor[rgb]{ .741,  .843,  .933} \textbf{93.8} & \cellcolor[rgb]{ .886,  .937,  .855} 59.0  & \cellcolor[rgb]{ .886,  .937,  .855} 66.7  & \cellcolor[rgb]{ .886,  .937,  .855} 77.5  \\
\midrule
\multirow{6}[2]{*}{EVA02} & Full  & 304.20M & 90.2  & 68.8  & 81.0  & 53.7  & 49.9  & 48.1  & 68.7  & 64.2  & 80.1  & 57.4  & 88.1  & 68.8  & 41.8  & 89.7  & 74.1  & 82.1  & 89.7  & 50.0  & 56.8  & 68.6  \\
      & Freeze & 0M    & 86.0  & 60.5  & 76.3  & 49.0  & 41.7  & 46.1  & 60.5  & 61.0  & 72.1  & 49.8  & 77.7  & 56.7  & 40.6  & 80.3  & 68.3  & 77.2  & 85.5  & 46.7  & 56.4  & 62.8  \\
      & REIN \cite{Wei_2024_CVPR} & 2.99M & 88.7  & \cellcolor[rgb]{ .886,  .937,  .855} 71.8  & \cellcolor[rgb]{ .886,  .937,  .855} 81.7  & \cellcolor[rgb]{ .886,  .937,  .855} 55.2  & 51.7  & \cellcolor[rgb]{ .886,  .937,  .855} 50.5  & \cellcolor[rgb]{ .886,  .937,  .855} 70.5  & \cellcolor[rgb]{ .886,  .937,  .855} 64.9  & \cellcolor[rgb]{ .886,  .937,  .855} 83.7  & 59.0  & \cellcolor[rgb]{ .741,  .843,  .933} \textbf{90.3} & \cellcolor[rgb]{ .741,  .843,  .933} \textbf{72.0} & \cellcolor[rgb]{ .886,  .937,  .855} 48.3  & \cellcolor[rgb]{ .741,  .843,  .933} \textbf{93.0} & \cellcolor[rgb]{ .886,  .937,  .855} 79.3  & \cellcolor[rgb]{ .886,  .937,  .855} 83.3  & \cellcolor[rgb]{ .886,  .937,  .855} 91.3  & 50.8  & \cellcolor[rgb]{ .886,  .937,  .855} 62.0  & \cellcolor[rgb]{ .886,  .937,  .855} 70.9  \\
      & VQT \cite{tu2023visual} & 3.01M & \cellcolor[rgb]{ .886,  .937,  .855} 90.3  & 71.2  & 81.4  & 54.3  & \cellcolor[rgb]{ .886,  .937,  .855} 53.1  & 49.1  & 67.9  & 64.3  & 82.0  & \cellcolor[rgb]{ .741,  .843,  .933} \textbf{60.5} & 86.9  & 66.8  & 41.3  & 89.3  & 76.6  & 81.7  & 91.3  & 47.2  & 55.7  & 69.0  \\
      & ChildTune \cite{xu2021raisechildlargelanguage} & 15.21M & 86.4  & 68.8  & 81.0  & 54.4  & 50.6  & 48.9  & 69.6  & 64.5  & 83.2  & 57.8  & 88.2  & 69.0  & 47.9  & 90.2  & 74.8  & 82.8  & 90.3  & \cellcolor[rgb]{ .886,  .937,  .855} 51.0  & 61.4  & 69.5  \\
      & Ours  & 15.21M & \cellcolor[rgb]{ .741,  .843,  .933} \textbf{90.5} & \cellcolor[rgb]{ .741,  .843,  .933} \textbf{75.2} & \cellcolor[rgb]{ .741,  .843,  .933} \textbf{83.6} & \cellcolor[rgb]{ .741,  .843,  .933} \textbf{58.8} & \cellcolor[rgb]{ .741,  .843,  .933} \textbf{54.6} & \cellcolor[rgb]{ .741,  .843,  .933} \textbf{52.2} & \cellcolor[rgb]{ .741,  .843,  .933} \textbf{73.1} & \cellcolor[rgb]{ .741,  .843,  .933} \textbf{66.6} & \cellcolor[rgb]{ .741,  .843,  .933} \textbf{85.7} & \cellcolor[rgb]{ .886,  .937,  .855} 60.5  & \cellcolor[rgb]{ .886,  .937,  .855} 90.2  & \cellcolor[rgb]{ .886,  .937,  .855} 70.7  & \cellcolor[rgb]{ .741,  .843,  .933} \textbf{51.5} & \cellcolor[rgb]{ .886,  .937,  .855} 92.3  & \cellcolor[rgb]{ .741,  .843,  .933} \textbf{82.6} & \cellcolor[rgb]{ .741,  .843,  .933} \textbf{88.2} & \cellcolor[rgb]{ .741,  .843,  .933} \textbf{91.9} & \cellcolor[rgb]{ .741,  .843,  .933} \textbf{54.0} & \cellcolor[rgb]{ .741,  .843,  .933} \textbf{62.4} & \cellcolor[rgb]{ .741,  .843,  .933} \textbf{72.9} \\
\bottomrule
\end{tabular}%

}
  \setlength{\abovecaptionskip}{0.05 cm}
  \caption{DGSS generalization performance for each category from the Cityscapes source domain to mixed-domain BDD100K and ACDC, with comparison methods including adaptor-based Rein \cite{Wei_2024_CVPR} and selective parameter fine-tuning methods VQT \cite{tu2023visual} and ChildTune\cite{xu2021raisechildlargelanguage}.}
  \label{exp_vfm_bdd}
  \vspace{-0.3cm}
\end{table*}

\begin{table*}[tbp]
  \centering
   \renewcommand{\arraystretch}{1.05}
  \resizebox{\linewidth}{!}{
\begin{tabular}{cccccccccc}
\toprule
      & \multicolumn{8}{c}{Cityscapes → Adverse Weather}              &  \\
\midrule
      & Fine-tune Method & Trainable Params & Foggy Zurich \cite{Sakaridis_2018_CVPR}  & Foggy Driving \cite{Sakaridis_2018_CVPR} & Dark Zurich \cite{sakaridis2019guided} & Nighttime Driving \cite{sakaridis2021nighttime} & ACDC-Rain \cite{sakaridis2021acdc}  & ACDC-Snow \cite{sakaridis2021acdc} & mIoU \\
\midrule
\multirow{5}[2]{*}{\makecell{ DINOv2 \cite{dinov2} \\
(Large)}} & Full  & 304.20M & 50.4  & 55.3  & 62.7  & 47.7  & 75.2  & 76.8  & 61.3  \\
      & Freeze & 0M    & 50.3  & 43.7  & 54.3  & 40.8  & 66.1  & 71.7  & 54.5  \\
      & REIN \cite{Wei_2024_CVPR} & 2.99M & \cellcolor[rgb]{ .886,  .937,  .855}55.5  & \cellcolor[rgb]{ .886,  .937,  .855}58.2  & 64.3  & 50.3  & \cellcolor[rgb]{ .886,  .937,  .855}78.2  & \cellcolor[rgb]{ .886,  .937,  .855}79.5  & \cellcolor[rgb]{ .886,  .937,  .855}64.3  \\
      & VQT \cite{tu2023visual}  & 3.01M  & 54.1  & 57.1  & 61.9  & 47.4  & 76.1  & 75.3  & 62.0  \\
      & ChildTune \cite{xu2021raisechildlargelanguage} & 15.21M & 55.2  & 56.9  & \cellcolor[rgb]{ .886,  .937,  .855}64.5  & \cellcolor[rgb]{ .886,  .937,  .855}50.7  & 77.7  & 78.3  & 63.9  \\
      & Ours & 15.21M & \cellcolor[rgb]{ .741,  .843,  .933} \textbf{56.9}  & \cellcolor[rgb]{ .741,  .843,  .933} \textbf{60.0}  & \cellcolor[rgb]{ .741,  .843,  .933} \textbf{66.6}  &\cellcolor[rgb]{ .741,  .843,  .933} \textbf{53.2}  & \cellcolor[rgb]{ .741,  .843,  .933} \textbf{78.6}  & \cellcolor[rgb]{ .741,  .843,  .933} \textbf{82.2}  & \cellcolor[rgb]{ .741,  .843,  .933} \textbf{66.3}  \\
\bottomrule
\end{tabular}%
}
 \setlength{\abovecaptionskip}{0.05 cm}
  \caption{DGSS performance comparison for Cityscapes as the source domain under diverse weather conditions.}
  \label{exp_vfm_aw}
  \vspace{-0.5cm}
\end{table*}

\subsection{Comparison with State-of-the-art Alternatives}

\begin{table}[t]
  \centering
   \renewcommand{\arraystretch}{1.05}
  \resizebox{\linewidth}{!}{
\begin{tabular}{cccc}
\toprule
      &       & Cityscapes →BDD100K & Cityscapes →ACDC \\
\midrule
\multirow{9}[4]{*}{\makecell{ EVA02 \cite{eva02} \\
(Large)}} & Full  & 62.1  & 68.6  \\
      & Freeze & 56.5  & 62.8  \\
\cmidrule{2-4}      &  Random  & 61.1  & 67.6  \\
      &   Random  $Q$ & 62.8  & 69.1  \\
      &   Random  $K$ & 61.9  & 68.1  \\
      &   Random  $V$ & 62.9  & 69.2  \\
      &   $  \mathbf{F}_{\boldsymbol{\theta}}$ & \cellcolor[rgb]{ .886,  .937,  .855} 63.8  & 69.5  \\
      &   $\Delta \mathbf{F}_{\boldsymbol{\theta}}$& 63.1  & \cellcolor[rgb]{ .886,  .937,  .855} 71.3  \\
      &   $\mathbf{DRF}_{\boldsymbol{\theta}}$ & \cellcolor[rgb]{ .741,  .843,  .933} \textbf{65.8 (+3.7)} & \cellcolor[rgb]{ .741,  .843,  .933} \textbf{72.9 (+5.3)} \\
\midrule
\multirow{9}[4]{*}{\makecell{ DINOv2 \cite{dinov2} \\
(Large)}} & Full  & 63.7  & 71.7  \\
      & Freeze & 63.3  & 67.5  \\
\cmidrule{2-4}      & Random  & 62.7  & 71.0  \\
      &   Random  $Q$ & 63.2  & 72.0  \\
      &   Random  $K$ & 63.5  & 72.3  \\
      &   Random  $V$ & 63.2  & 72.9  \\
      &   $  \mathbf{F}_{\boldsymbol{\theta}}$ & 63.8  & 71.4  \\
      &   $\Delta \mathbf{F}_{\boldsymbol{\theta}}$ & \cellcolor[rgb]{ .886,  .937,  .855} 64.5  & \cellcolor[rgb]{ .886,  .937,  .855} 76.1  \\
      &  $\mathbf{DRF}_{\boldsymbol{\theta}}$ & \cellcolor[rgb]{ .741,  .843,  .933} \textbf{67.7 (+4.0)} & \cellcolor[rgb]{ .741,  .843,  .933} \textbf{77.5 (+5.8)} \\
\bottomrule
\end{tabular}%
}
 \setlength{\abovecaptionskip}{0.05 cm}
  \caption{Ablation study on generalization with 5\% fine-tunable parameters in terms of mIoU.}
  \label{exp_Ablation}
  \vspace{-0.5cm}
\end{table}

\noindent \textbf{GTAV → {C, B, M}.}
Table~\ref{exp_vfm_all} demonstrates that our approach significantly outperforms other fine-tuning methods across multiple vision foundation models (VFMs). Compared to adapter-based methods (e.g., LoRA and Rein), our approach achieves an average of 4.3\% higher mIoU than Rein across five VFM models. Additionally, it surpasses the self-focused parametric fine-tuning method VQT by 3.1\% on average. Notably, for models with a substantial gap between pre-training and downstream tasks, such as MAE and EVA02, adapter methods yielded modest improvements of 1.3\% and 1.7\% mIoU, respectively, whereas our approach achieved 4.6\% and 6.6\% improvements. 
Besides, we added comparisons with the state-of-the-art methods using VFMs, and our method remains competitive. The tqdm\cite{pak2024textual} and VLTSeg\cite{hummer2024strong} method leverages features of the language model, while Rein-series methods and ours focus on visual models.
These results highlight our method’s enhanced adaptability to downstream tasks and its significant improvement in model generalization.

\noindent \textbf{Cityscapes → {BDD100K, ACDC}.}
In migrating from Cityscapes to BDD100K and ACDC, our method achieved strong results, with average mIoU scores of 67.7\% and 77.5\%. As shown in Table~\ref{exp_vfm_bdd}, our method’s average mIoU on BDD100K is 2.4\% higher than REIN. Compared to VQT and ChildTune, our method improved mIoU by 4.4\% and 2.6\% on the respective datasets, addressing issues in parameter tuning, data adaptation, and migration strategy. These results highlight our method’s superior adaptability and generalization in complex scenes.

\noindent \textbf{Cityscapes → {Adverse Weather}.}
We evaluated various fine-tuning strategies for DINOv2 models across challenging weather conditions, as shown in Table~\ref{exp_vfm_aw}.
Our approach achieved an average of 2.0\% higher mIoU than the adapter-based REIN method and 4.3\% higher than the self-focused VQT approach. This improvement likely stems from the substantial difference between pre-training and downstream tasks. Besides, ChildTune showed limited performance gains, and our method surpassed ChildTune by an average of 2.4\% mIoU, demonstrating superior adaptability and generalization under complex weather scenarios.

\vspace{-0.2cm}
\subsection{Ablation Studies}
\vspace{-0.2cm}

\noindent \textbf{Ablation of DR-FIM effectiveness} As shown in Table~\ref{exp_Ablation}, randomly selecting $Q$, $K$, and $V$ parameters for fine-tuning does not fully leverage the generalization ability of VFMs, leading to lower mIoU. Using FIM ($\mathbf{F}_{\boldsymbol{\theta}}$) for parameter selection improves performance over random choice. Further gains are achieved with $\Delta \mathbf{F}_{\boldsymbol{\theta}}$, which better identifies domain-sensitive parameters—especially on ACDC, where severe weather differences pose greater challenges. Our proposed DR-FIM, combining \(F\) and \(\Delta F\), delivers the best results, boosting mIoU by +3.7\% and +5.3\% on Cityscapes → BDD100K and Cityscapes → ACDC for EVA02 (Large), and by +4.0\% and +5.8\%, respectively. These results highlight the effectiveness of our method.

\begin{figure}[t]
    \begin{center}
    \centering 
    \includegraphics[width=0.425\textwidth]{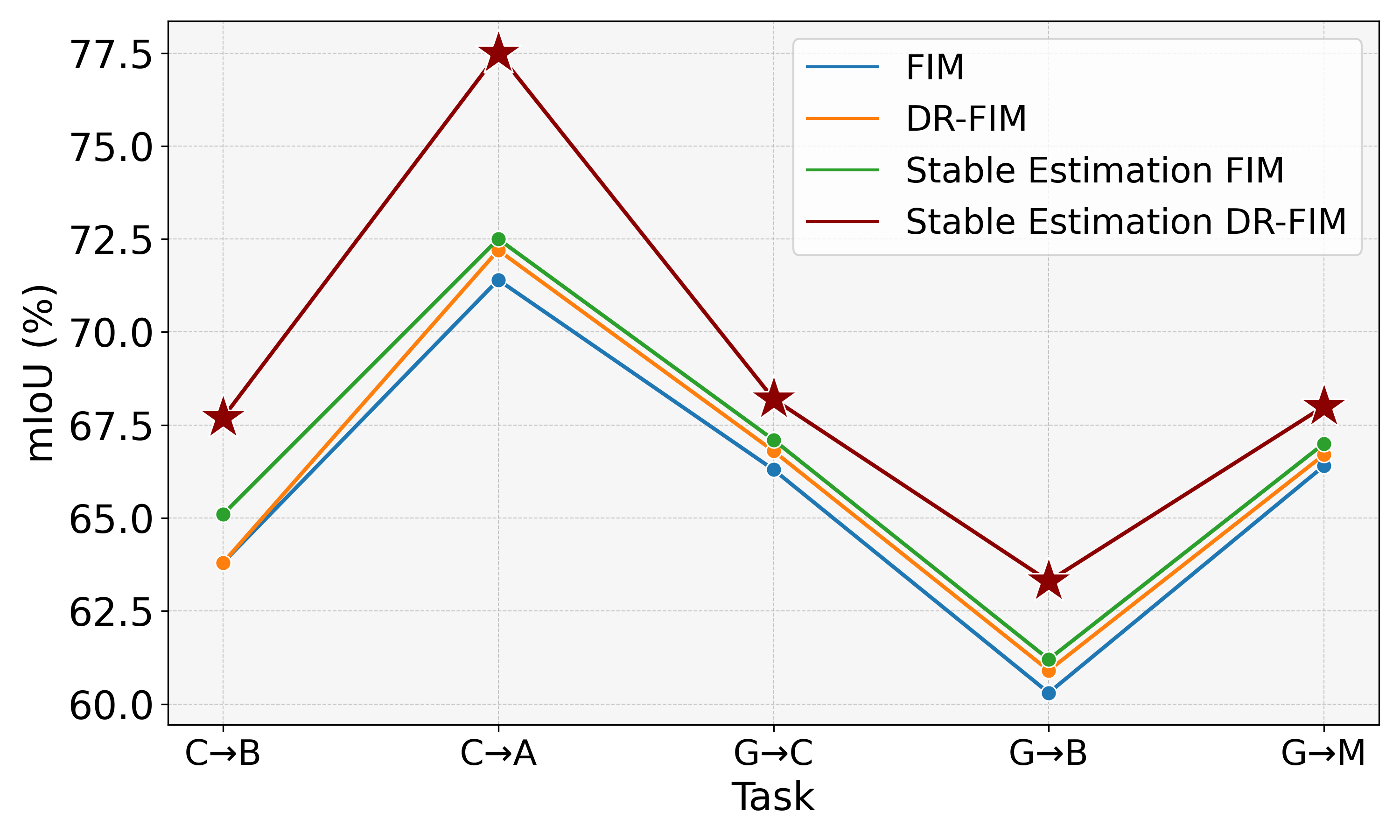} 
    \end{center} 
    \setlength{\abovecaptionskip}{-0.5cm}
    \caption{Ablation study of estimation ways on Cityscapes → BDD100K (C2B), →ACDC (C2A), and GTAV → Cityscapes(G2C), → BDD100K(G2B) and  → Mapillary(G2M).}
    \label{alb_stable}
\vspace{-0.5cm}
\end{figure}

\noindent \textbf{Ablation of DR-FIM Estimation}
Fig.~\ref{alb_stable} presents the ablation study on the proposed stable estimation method. The results show that while DR-FIM outperforms FIM in parameter evaluation, but the effectiveness of DR-FIM is limited by traditional estimation methods. The stable estimation method significantly enhances the accuracy of parameter evaluation for both FIM and DR-FIM. Notably, applying stable estimation to DR-FIM results in an average improvement of 2.6\% mIoU, demonstrating superior overall generalization performance.

\noindent \textbf{Ablation of Feature Perturbation} Since we adopt domain simulation augmentation from \cite{li2021uncertainty}, which is generally considered effective for DG, we also apply it to existing VFM methods for a fair comparison. FisherTune uses feature perturbation (FP) solely for identifying domain-sensitive parameters, not during fine-tuning. As shown in Table~\ref{alb_FP}, FP yields a modest improvement (+1.0\% mIoU) on GTA→Avg., yet our method still outperforms others.

\vspace{-0.15cm}
\subsection{Discussion}
\vspace{-0.15cm}

\noindent \textbf{Captured domain-sensitive parameters.}
Fig.~\ref{qkv_show} illustrates the impact of different estimation methods on parameter sensitivity estimation. 
(a) shows that parameter sensitivity estimated from original FIM is generally high, making it difficult to identify the most valuable parameters.  
(b) demonstrates that incorporating \(\Delta \mathbf{F}_{\boldsymbol{\theta}}\) redefines parameter sensitivity by comprehensively considering both task relevance and domain sensitivity.  
(c) presents the DR-FIM estimated using a robust way, which highlights important parameters more effectively, aiding in the selection of valuable parameters.  
Additionally, (c) reveals that important parameters tend to be concentrated in the $Q$, $K$, $V$ and FFM parameters of deeper blocks. Furthermore, the overall sensitivity of \(Q\) and \(K\) is higher than that of \(V\).

\noindent \textbf{Feature Visualization.} 
Fig.~\ref{tsne} compares the T-SNE visualizations of feature distributions between Rein\cite{Wei_2024_CVPR} and FisherTune. FisherTune exhibits a more balanced feature distribution across multiple unseen domains, indicating reduced domain bias and improved generalization.

\noindent \textbf{The Ratio of Fine-tuned Parameters}. \textcolor[rgb]{0,0.69,0.94}{\textbf{See Appendix C}}.

\noindent \textbf{Segmentation Result Visualization}. \textcolor[rgb]{0,0.69,0.94}{\textbf{See Appendix D}}.

\noindent \textbf{Influence of Hyper-parameters}. \textcolor[rgb]{0,0.69,0.94}{\textbf{See Appendix E}}.

\begin{table}[t]
\vspace{-0.1cm}
  \centering
  \resizebox{0.95\linewidth}{!}{
\begin{tabular}{ccccc}
\toprule
Method & EVA02 & EVA02+FP & DINOV2 & DINOV2+FP \\
\midrule
AdaptFormer & 62.6  & 63.3 (+0.7) & 62.7  & 63.7 (+1.0) \\
VPT   & 60.8  & 61.8 (+1.0) & 63.3  & 64.1 (+0.8) \\
Rein  & 63.6  & 63.9 (+0.3) & 64.3  & 65.0 (+0.7) \\
\rowcolor[rgb]{ .867,  .922,  .969} Ours & 64.4  & 64.5 (+0.1) & 66.3  & 66.5 (+0.2) \\
\bottomrule
\end{tabular}%
}
 \setlength{\abovecaptionskip}{0.05 cm}
  \caption{Ablation study on Feature Perturbation (FP) using~\cite{li2022uncertainty}.}
  \label{alb_FP}
\vspace{-0.6cm}
\end{table}

\begin{figure}[t]
    \begin{center}
    \centering 
    \includegraphics[width=0.45\textwidth]{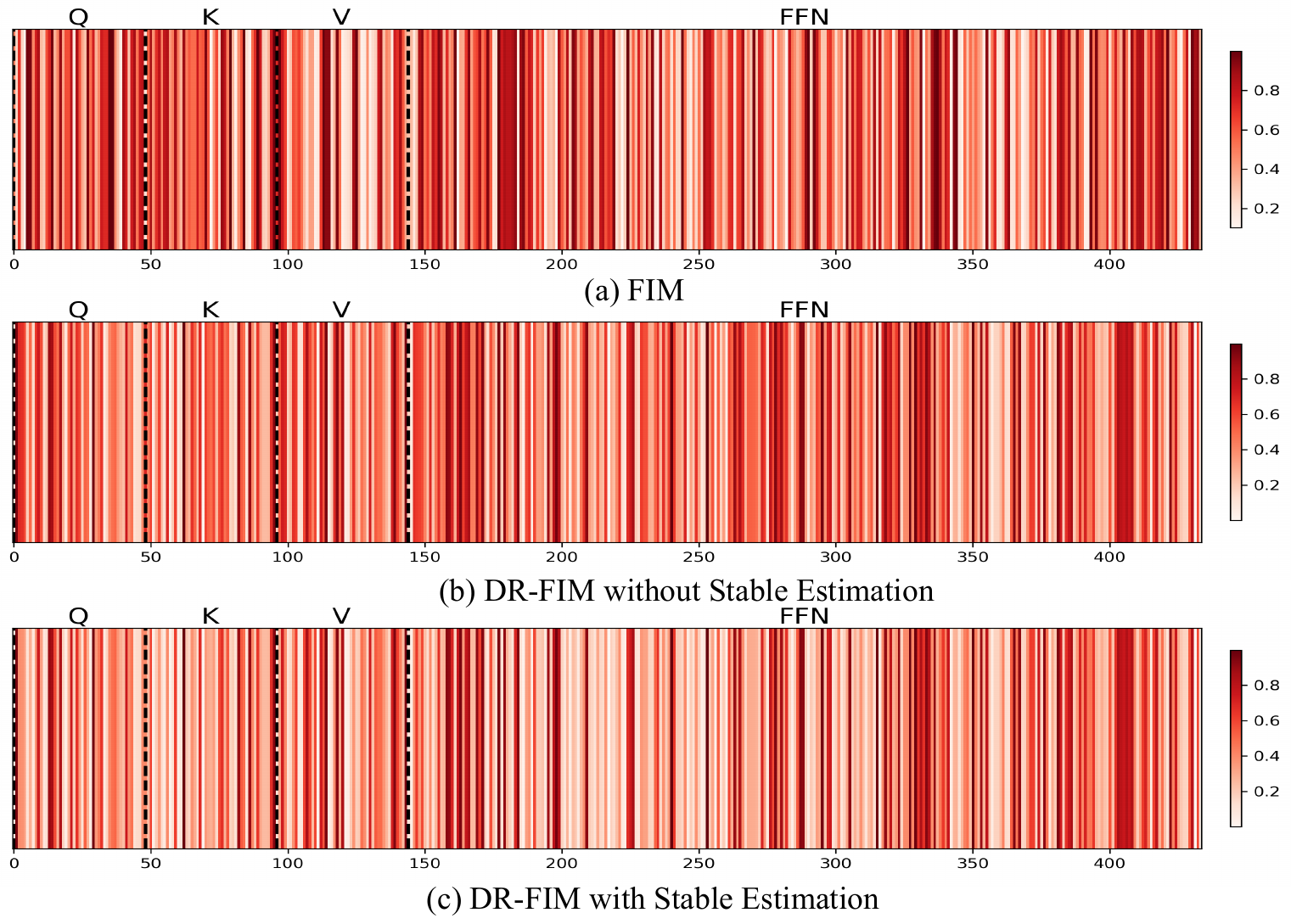} 
    \end{center} 
    \setlength{\abovecaptionskip}{-0.4cm}
    \caption{Diagram of parameter sensitivity estimated by FIM and our DR-FIM using DINOV2-large, trained on GTAV for DGSS experiments. The $Q$, $K$, $V$, and FFN parameters are arranged in ascending order according to their block indices.} 
    \label{qkv_show}
\vspace{-0.4cm}
\end{figure}

\begin{figure}[t]
    \begin{center}
    \centering 
    \includegraphics[width=0.45\textwidth]{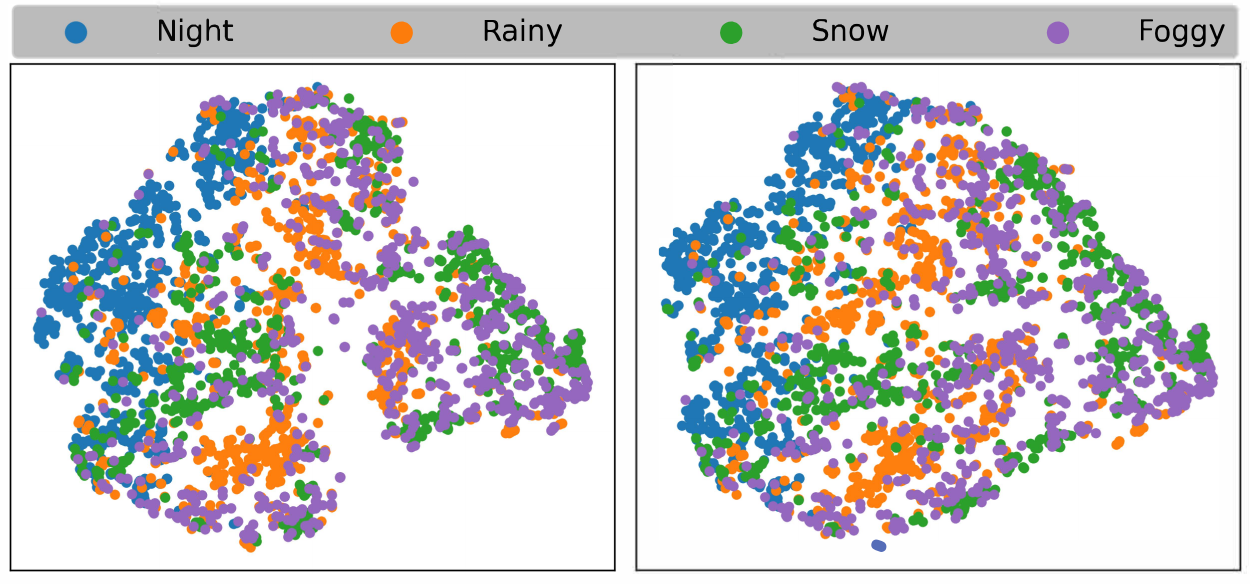} 
    \end{center} 
    \setlength{\abovecaptionskip}{-0.3cm}
    \caption{Comparison of T-SNE feature visualizations: Rein\cite{Wei_2024_CVPR} (left) and the proposed FisherTune (right). The model is trained on the Cityscapes → ACDC DGSS task. FisherTune shows a more balanced feature distribution across multiple unseen domains.}
    \label{tsne}
\vspace{-0.3cm}
\end{figure}

\vspace{-0.2cm}
\section{Conclusion}
\vspace{-0.2cm}
We propose FisherTune, a fine-tuning method for Vision Foundation Models (VFMs) in DGSS. It introduces the Domain-Related Fisher Information Matrix (DR-FIM) to measure parameter sensitivity to domain shifts, using variational inference for stable estimation. FisherTune enhances domain adaptability while maintaining generalization. We hope it encourages further research on selective fine-tuning to better unlock the generalization potential of VFMs in DGSS and beyond.

{\small
\bibliographystyle{ieee_fullname}
\bibliography{ref}

\begin{thebibliography}{10}\itemsep=-1pt

\bibitem{achille2019task2vec}
Alessandro Achille, Michael Lam, Rahul Tewari, Avinash Ravichandran, Subhransu
  Maji, Charless~C Fowlkes, Stefano Soatto, and Pietro Perona.
\newblock Task2vec: Task embedding for meta-learning.
\newblock In {\em Proceedings of the IEEE/CVF international conference on
  computer vision}, pages 6430--6439, 2019.

\bibitem{achille2019information}
Alessandro Achille, Giovanni Paolini, and Stefano Soatto.
\newblock Where is the information in a deep neural network?
\newblock {\em arXiv preprint arXiv:1905.12213}, 2019.

\bibitem{benigmim2024collaborating}
Yasser Benigmim, Subhankar Roy, Slim Essid, Vicky Kalogeiton, and St{\'e}phane
  Lathuili{\`e}re.
\newblock Collaborating foundation models for domain generalized semantic
  segmentation.
\newblock In {\em Proceedings of the IEEE/CVF Conference on Computer Vision and
  Pattern Recognition}, pages 3108--3119, 2024.

\bibitem{blei2017variational}
David~M Blei, Alp Kucukelbir, and Jon~D McAuliffe.
\newblock Variational inference: A review for statisticians.
\newblock {\em Journal of the American statistical Association},
  112(518):859--877, 2017.

\bibitem{dinov2}
Mathilde Caron et~al.
\newblock Dinov2: Learning robust visual features without supervision.
\newblock {\em arXiv preprint arXiv:2304.07193}, 2023.

\bibitem{chattopadhyay2023pasta}
Prithvijit Chattopadhyay, Kartik Sarangmath, Vivek Vijaykumar, and Judy
  Hoffman.
\newblock Pasta: Proportional amplitude spectrum training augmentation for
  syn-to-real domain generalization.
\newblock In {\em Proceedings of the IEEE/CVF International Conference on
  Computer Vision}, pages 19288--19300, 2023.

\bibitem{chen2022adaptformer}
Shoufa Chen, Chongjian Ge, Zhan Tong, Jiangliu Wang, Yibing Song, Jue Wang, and
  Ping Luo.
\newblock Adaptformer: Adapting vision transformers for scalable visual
  recognition.
\newblock {\em Advances in Neural Information Processing Systems},
  35:16664--16678, 2022.

\bibitem{cheng2022domain}
Ziyuan Cheng, Ruinian Wan, Meng Li, Feiyu Wang, Chao Xu, and Xiaofei He.
\newblock Domain generalization via style-efficient perturbation and clustering
  of intra-domain heterogeneous data.
\newblock In {\em Proceedings of the IEEE/CVF Conference on Computer Vision and
  Pattern Recognition}, pages 3938--3947, 2022.

\bibitem{choi2021robustnet}
Sungha Choi, Sanghun Jung, Huiwon Yun, Joanne~T Kim, Seungryong Kim, and Jaegul
  Choo.
\newblock Robustnet: Improving domain generalization in urban-scene
  segmentation via instance selective whitening.
\newblock In {\em Proceedings of the IEEE/CVF conference on computer vision and
  pattern recognition}, pages 11580--11590, 2021.

\bibitem{dosovitskiy2020image}
Alexey Dosovitskiy.
\newblock An image is worth 16x16 words: Transformers for image recognition at
  scale.
\newblock {\em arXiv preprint arXiv:2010.11929}, 2020.

\bibitem{dou2019domain}
Qiong Dou, Daniel~Caro de Castro, Konstantinos Kamnitsas, and Ben Glocker.
\newblock Domain generalization via model-agnostic learning of semantic
  features.
\newblock In {\em Advances in Neural Information Processing Systems}, pages
  6450--6461, 2019.

\bibitem{fahes2023poda}
Mohammad Fahes, Tuan-Hung Vu, Andrei Bursuc, Patrick P{\'e}rez, and Raoul
  De~Charette.
\newblock Poda: Prompt-driven zero-shot domain adaptation.
\newblock In {\em Proceedings of the IEEE/CVF International Conference on
  Computer Vision}, pages 18623--18633, 2023.

\bibitem{fahes2024simple}
Mohammad Fahes, Tuan-Hung Vu, Andrei Bursuc, Patrick P{\'e}rez, and Raoul de
  Charette.
\newblock A simple recipe for language-guided domain generalized segmentation.
\newblock In {\em Proceedings of the IEEE/CVF Conference on Computer Vision and
  Pattern Recognition}, pages 23428--23437, 2024.

\bibitem{fang2023eva02}
Hao Fang et~al.
\newblock Eva-02: A visual learner for more generalized visual representation
  learning.
\newblock In {\em Conference on Computer Vision and Pattern Recognition
  (CVPR)}, 2023.

\bibitem{eva02}
Hao Fang et~al.
\newblock Eva-clip: Improving vision-language models with masked modeling.
\newblock {\em arXiv preprint arXiv:2303.13495}, 2023.

\bibitem{fisher1922mathematical}
Ronald~A Fisher.
\newblock On the mathematical foundations of theoretical statistics.
\newblock {\em Philosophical transactions of the Royal Society of London.
  Series A, containing papers of a mathematical or physical character},
  222(594-604):309--368, 1922.

\bibitem{gao2024mini}
Zhangwei Gao, Zhe Chen, Erfei Cui, Yiming Ren, Weiyun Wang, Jinguo Zhu, Hao
  Tian, Shenglong Ye, Junjun He, Xizhou Zhu, et~al.
\newblock Mini-internvl: a flexible-transfer pocket multi-modal model with 5\%
  parameters and 90\% performance.
\newblock {\em Visual Intelligence}, 2(1):1--17, 2024.

\bibitem{han2024parameter}
Zeyu Han, Chao Gao, Jinyang Liu, Jeff Zhang, and Sai~Qian Zhang.
\newblock Parameter-efficient fine-tuning for large models: A comprehensive
  survey.
\newblock {\em arXiv preprint arXiv:2403.14608}, 2024.

\bibitem{he2022mae}
Kaiming He et~al.
\newblock Masked autoencoders are scalable vision learners.
\newblock In {\em Conference on Computer Vision and Pattern Recognition
  (CVPR)}, 2022.

\bibitem{hoffman2013stochastic}
Matthew~D Hoffman, David~M Blei, Chong Wang, and John Paisley.
\newblock Stochastic variational inference.
\newblock {\em Journal of Machine Learning Research}, 2013.

\bibitem{lora}
Edward~J Hu et~al.
\newblock Lora: Low-rank adaptation of large language models.
\newblock {\em International Conference on Learning Representations (ICLR)},
  2022.

\bibitem{hu2021lora}
Edward~J Hu, Yelong Shen, Phillip Wallis, Zeyuan Allen-Zhu, Yuanzhi Li, Shean
  Wang, Lu Wang, and Weizhu Chen.
\newblock Lora: Low-rank adaptation of large language models.
\newblock {\em arXiv preprint arXiv:2106.09685}, 2021.

\bibitem{huang2021fsdr}
Jiaxing Huang, Dayan Guan, Aoran Xiao, and Shijian Lu.
\newblock Fsdr: Frequency space domain randomization for domain generalization.
\newblock In {\em Proceedings of the IEEE/CVF conference on computer vision and
  pattern recognition}, pages 6891--6902, 2021.

\bibitem{hummer2024strong}
Christoph H{\"u}mmer, Manuel Schwonberg, Liangwei Zhou, Hu Cao, Alois Knoll,
  and Hanno Gottschalk.
\newblock Strong but simple: A baseline for domain generalized dense perception
  by clip-based transfer learning.
\newblock In {\em Proceedings of the Asian Conference on Computer Vision},
  pages 4223--4244, 2024.

\bibitem{jia2022visual}
Menglin Jia, Luming Tang, Bor-Chun Chen, Claire Cardie, Serge Belongie, Bharath
  Hariharan, and Ser-Nam Lim.
\newblock Visual prompt tuning.
\newblock In {\em European Conference on Computer Vision}, pages 709--727.
  Springer, 2022.

\bibitem{khattak2023maple}
Muhammad~Uzair Khattak, Hanoona Rasheed, Muhammad Maaz, Salman Khan, and
  Fahad~Shahbaz Khan.
\newblock Maple: Multi-modal prompt learning.
\newblock In {\em Proceedings of the IEEE/CVF Conference on Computer Vision and
  Pattern Recognition}, pages 19113--19122, 2023.

\bibitem{kingma2015variational}
Durk~P Kingma, Tim Salimans, and Max Welling.
\newblock Variational dropout and the local reparameterization trick.
\newblock {\em Advances in neural information processing systems}, 28, 2015.

\bibitem{kirillov2023sam}
Alexander Kirillov et~al.
\newblock Segment anything.
\newblock {\em arXiv preprint arXiv:2304.02643}, 2023.

\bibitem{lee2022wildnet}
Suhyeon Lee, Hongje Seong, Seongwon Lee, and Euntai Kim.
\newblock Wildnet: Learning domain generalized semantic segmentation from the
  wild.
\newblock In {\em Proceedings of the IEEE/CVF conference on computer vision and
  pattern recognition}, pages 9936--9946, 2022.

\bibitem{li2018learning}
Da Li, Yongxin Yang, Yi-Zhe Song, and Timothy~M Hospedales.
\newblock Learning to generalize: Meta-learning for domain generalization.
\newblock In {\em Proceedings of the AAAI Conference on Artificial
  Intelligence}, volume~32, 2018.

\bibitem{li2023blip}
Junnan Li, Dongxu Li, Silvio Savarese, and Steven Hoi.
\newblock Blip-2: Bootstrapping language-image pre-training with frozen image
  encoders and large language models.
\newblock In {\em International conference on machine learning}, pages
  19730--19742. PMLR, 2023.

\bibitem{li2021uncertainty}
Wei Li, Guoqiang Niu, and Wen Liu.
\newblock Uncertainty-aware unsupervised domain adaptation for semantic
  segmentation.
\newblock {\em Neurocomputing}, 419:103--115, 2021.

\bibitem{li2022uncertainty}
Xiaotong Li, Yongxing Dai, Yixiao Ge, Jun Liu, Ying Shan, and Ling-Yu Duan.
\newblock Uncertainty modeling for out-of-distribution generalization.
\newblock {\em arXiv preprint arXiv:2202.03958}, 2022.

\bibitem{liao2023parameterefficientfinetuningintroducingnew}
Baohao Liao, Yan Meng, and Christof Monz.
\newblock Parameter-efficient fine-tuning without introducing new latency,
  2023.

\bibitem{liu2023explicit}
Weihuang Liu, Xi Shen, Chi-Man Pun, and Xiaodong Cun.
\newblock Explicit visual prompting for low-level structure segmentations.
\newblock In {\em Proceedings of the IEEE/CVF Conference on Computer Vision and
  Pattern Recognition}, pages 19434--19445, 2023.

\bibitem{matena2022merging}
Michael~S Matena and Colin~A Raffel.
\newblock Merging models with fisher-weighted averaging.
\newblock {\em Advances in Neural Information Processing Systems},
  35:17703--17716, 2022.

\bibitem{moor2023foundation}
Michael Moor, Oishi Banerjee, Zahra Shakeri~Hossein Abad, Harlan~M Krumholz,
  Jure Leskovec, Eric~J Topol, and Pranav Rajpurkar.
\newblock Foundation models for generalist medical artificial intelligence.
\newblock {\em Nature}, 616(7956):259--265, 2023.

\bibitem{murez2018image}
Zak Murez, Nicholas Kolkin, Noriaki Liu, Karttikeya Ramachandran, and Alexei~A
  Efros.
\newblock Image to image translation for domain adaptation.
\newblock In {\em Proceedings of the IEEE/CVF Conference on Computer Vision and
  Pattern Recognition}, pages 4500--4509, 2018.

\bibitem{nam2021reducing}
Hyeonseong Nam, Jaeyeon Kim, Donggeun Kim, and Bohyung Han.
\newblock Reducing domain gap via style-agnostic networks.
\newblock In {\em Proceedings of the IEEE/CVF Conference on Computer Vision and
  Pattern Recognition}, pages 8690--8700, 2021.

\bibitem{niemeijer2024generalization}
Joshua Niemeijer, Manuel Schwonberg, Jan-Aike Term{\"o}hlen, Nico~M Schmidt,
  and Tim Fingscheidt.
\newblock Generalization by adaptation: Diffusion-based domain extension for
  domain-generalized semantic segmentation.
\newblock In {\em Proceedings of the IEEE/CVF Winter Conference on Applications
  of Computer Vision}, pages 2830--2840, 2024.

\bibitem{oquab2023dinov2}
Maxime Oquab et~al.
\newblock Dinov2: Learning robust visual features without supervision.
\newblock {\em arXiv preprint arXiv:2304.07193}, 2023.

\bibitem{pak2024textual}
Byeonghyun Pak, Byeongju Woo, Sunghwan Kim, Dae-hwan Kim, and Hoseong Kim.
\newblock Textual query-driven mask transformer for domain generalized
  segmentation.
\newblock In {\em European Conference on Computer Vision}, pages 37--54.
  Springer, 2024.

\bibitem{pan2022fourier}
Xingang Pan, Xiaohang Zhan, Bo Dai, and Ping Luo.
\newblock Fourier domain adaptation for semantic segmentation.
\newblock In {\em Proceedings of the IEEE/CVF Conference on Computer Vision and
  Pattern Recognition}, pages 15711--15721, 2022.

\bibitem{peng2022semantic}
Duo Peng, Yinjie Lei, Munawar Hayat, Yulan Guo, and Wen Li.
\newblock Semantic-aware domain generalized segmentation.
\newblock In {\em Proceedings of the IEEE/CVF conference on computer vision and
  pattern recognition}, pages 2594--2605, 2022.

\bibitem{peng2021global}
Duo Peng, Yinjie Lei, Lingqiao Liu, Pingping Zhang, and Jun Liu.
\newblock Global and local texture randomization for synthetic-to-real semantic
  segmentation.
\newblock {\em IEEE Transactions on Image Processing}, 30:6594--6608, 2021.

\bibitem{pu2020dual}
Nan Pu, Wei Chen, Yu Liu, Erwin~M Bakker, and Michael~S Lew.
\newblock Dual gaussian-based variational subspace disentanglement for
  visible-infrared person re-identification.
\newblock In {\em Proceedings of the 28th ACM international conference on
  multimedia}, pages 2149--2158, 2020.

\bibitem{pu2023dynamic}
Nan Pu, Zhun Zhong, and Nicu Sebe.
\newblock Dynamic conceptional contrastive learning for generalized category
  discovery.
\newblock In {\em Proceedings of the IEEE/CVF conference on computer vision and
  pattern recognition}, pages 7579--7588, 2023.

\bibitem{pu2023memorizing}
Nan Pu, Zhun Zhong, Nicu Sebe, and Michael~S Lew.
\newblock A memorizing and generalizing framework for lifelong person
  re-identification.
\newblock {\em IEEE Transactions on Pattern Analysis and Machine Intelligence},
  45(11):13567--13585, 2023.

\bibitem{clip}
Alec Radford, Jong~Wook Kim, Chris Hallacy, et~al.
\newblock Learning transferable visual models from natural language
  supervision.
\newblock {\em International Conference on Machine Learning (ICML)}, 2021.

\bibitem{radford2021learning}
Alec Radford, Jong~Wook Kim, Chris Hallacy, Aditya Ramesh, Gabriel Goh,
  Sandhini Agarwal, Girish Sastry, Amanda Askell, Pamela Mishkin, Jack Clark,
  et~al.
\newblock Learning transferable visual models from natural language
  supervision.
\newblock In {\em International conference on machine learning}, pages
  8748--8763. PMLR, 2021.

\bibitem{rame2022fishr}
Alexandre Rame, Corentin Dancette, and Matthieu Cord.
\newblock Fishr: Invariant gradient variances for out-of-distribution
  generalization.
\newblock In {\em International Conference on Machine Learning}, pages
  18347--18377. PMLR, 2022.

\bibitem{Sakaridis_2018_CVPR}
Christos Sakaridis, Dengxin Dai, and Luc Van~Gool.
\newblock Semantic foggy scene understanding with synthetic data.
\newblock In {\em Proceedings of the IEEE Conference on Computer Vision and
  Pattern Recognition (CVPR)}, June 2018.

\bibitem{sakaridis2019guided}
Christos Sakaridis, Dengxin Dai, and Luc Van~Gool.
\newblock Guided curriculum model adaptation and uncertainty-aware evaluation
  for semantic nighttime image segmentation.
\newblock In {\em Proceedings of the IEEE/CVF International Conference on
  Computer Vision}, pages 7374--7383, 2019.

\bibitem{sakaridis2021acdc}
Christos Sakaridis, Dengxin Dai, and Luc Van~Gool.
\newblock Acdc: The adverse conditions dataset with correspondences for
  semantic driving scene understanding.
\newblock In {\em Proceedings of the IEEE/CVF International Conference on
  Computer Vision}, pages 10765--10775, 2021.

\bibitem{sakaridis2021nighttime}
Christos Sakaridis, Dengxin Dai, and Luc Van~Gool.
\newblock Semantic nighttime image segmentation with synthetic stylized data,
  gradual adaptation and uncertainty-aware evaluation.
\newblock In {\em Proceedings of the IEEE/CVF Conference on Computer Vision and
  Pattern Recognition}, pages 5467--5477, 2021.

\bibitem{scheibenreif2024parameter}
Linus Scheibenreif, Michael Mommert, and Damian Borth.
\newblock Parameter efficient self-supervised geospatial domain adaptation.
\newblock In {\em Proceedings of the IEEE/CVF Conference on Computer Vision and
  Pattern Recognition}, pages 27841--27851, 2024.

\bibitem{shu2022test}
Manli Shu, Weili Nie, De-An Huang, Zhiding Yu, Tom Goldstein, Anima Anandkumar,
  and Chaowei Xiao.
\newblock Test-time prompt tuning for zero-shot generalization in
  vision-language models.
\newblock {\em Advances in Neural Information Processing Systems},
  35:14274--14289, 2022.

\bibitem{sung2021trainingneuralnetworksfixed}
Yi-Lin Sung, Varun Nair, and Colin Raffel.
\newblock Training neural networks with fixed sparse masks, 2021.

\bibitem{Tang_2024_CVPR}
Song Tang, Wenxin Su, Mao Ye, and Xiatian Zhu.
\newblock Source-free domain adaptation with frozen multimodal foundation
  model.
\newblock In {\em Proceedings of the IEEE/CVF Conference on Computer Vision and
  Pattern Recognition (CVPR)}, pages 23711--23720, June 2024.

\bibitem{tu2023visual}
Cheng-Hao Tu, Zheda Mai, and Wei-Lun Chao.
\newblock Visual query tuning: Towards effective usage of intermediate
  representations for parameter and memory efficient transfer learning.
\newblock In {\em Proceedings of the IEEE/CVF Conference on Computer Vision and
  Pattern Recognition}, pages 7725--7735, 2023.

\bibitem{9961139}
Shuang Wang, Dong Zhao, Chi Zhang, Yuwei Guo, Qi Zang, Yu Gu, Yi Li, and
  Licheng Jiao.
\newblock Cluster alignment with target knowledge mining for unsupervised
  domain adaptation semantic segmentation.
\newblock {\em IEEE Transactions on Image Processing}, 31:7403--7418, 2022.

\bibitem{wang2021learning}
Tianci Wang, Wenxin Zhu, Zhengguo Lu, and Xiaoguang Wang.
\newblock Learning robust representations for domain generalization in semantic
  segmentation.
\newblock In {\em Proceedings of the IEEE/CVF Conference on Computer Vision and
  Pattern Recognition}, pages 9535--9545, 2021.

\bibitem{Wei_2024_CVPR}
Zhixiang Wei, Lin Chen, Yi Jin, Xiaoxiao Ma, Tianle Liu, Pengyang Ling, Ben
  Wang, Huaian Chen, and Jinjin Zheng.
\newblock Stronger fewer \& superior: Harnessing vision foundation models for
  domain generalized semantic segmentation.
\newblock In {\em Proceedings of the IEEE/CVF Conference on Computer Vision and
  Pattern Recognition (CVPR)}, pages 28619--28630, June 2024.

\bibitem{xia2024unsupervised}
Ruihao Xia, Yu Liang, Peng-Tao Jiang, Hao Zhang, Bo Li, Yang Tang, and Pan
  Zhou.
\newblock Unsupervised modality adaptation with text-to-image diffusion models
  for semantic segmentation.
\newblock {\em arXiv preprint arXiv:2410.21708}, 2024.

\bibitem{xu2021raisechildlargelanguage}
Runxin Xu, Fuli Luo, Zhiyuan Zhang, Chuanqi Tan, Baobao Chang, Songfang Huang,
  and Fei Huang.
\newblock Raise a child in large language model: Towards effective and
  generalizable fine-tuning, 2021.

\bibitem{xu2021raise}
Runxin Xu, Fuli Luo, Zhiyuan Zhang, Chuanqi Tan, Baobao Chang, Songfang Huang,
  and Fei Huang.
\newblock Raise a child in large language model: Towards effective and
  generalizable fine-tuning.
\newblock {\em arXiv preprint arXiv:2109.05687}, 2021.

\bibitem{yang2024exploring}
Senqiao Yang, Jiarui Wu, Jiaming Liu, Xiaoqi Li, Qizhe Zhang, Mingjie Pan, Yulu
  Gan, Zehui Chen, and Shanghang Zhang.
\newblock Exploring sparse visual prompt for domain adaptive dense prediction.
\newblock In {\em Proceedings of the AAAI Conference on Artificial
  Intelligence}, volume~38, pages 16334--16342, 2024.

\bibitem{yao2023visual}
Hantao Yao, Rui Zhang, and Changsheng Xu.
\newblock Visual-language prompt tuning with knowledge-guided context
  optimization.
\newblock In {\em Proceedings of the IEEE/CVF conference on computer vision and
  pattern recognition}, pages 6757--6767, 2023.

\bibitem{yi2024learning}
Jingjun Yi, Qi Bi, Hao Zheng, Haolan Zhan, Wei Ji, Yawen Huang, Yuexiang Li,
  and Yefeng Zheng.
\newblock Learning spectral-decomposited tokens for domain generalized semantic
  segmentation.
\newblock In {\em Proceedings of the 32nd ACM International Conference on
  Multimedia}, pages 8159--8168, 2024.

\bibitem{yue2019domain}
Xiangyun Yue, Zhenzhen Wang, Rogerio Feris, Qixing Wang, and Hongyuan Zha.
\newblock Domain randomization and pyramid consistency: Simulation-to-real
  generalization without accessing target domain data.
\newblock In {\em Proceedings of the IEEE/CVF International Conference on
  Computer Vision}, pages 2100--2110, 2019.

\bibitem{zang_mm_GSF}
Qi Zang, Shuang Wang, Dong Zhao, Yang Hu, Dou Quan, Jinlong Li, Nicu Sebe, and
  Zhun Zhong.
\newblock Generalized source-free domain-adaptive segmentation via reliable
  knowledge propagation.
\newblock In {\em Proceedings of the 32nd ACM International Conference on
  Multimedia}, MM '24, page 5967–5976, New York, NY, USA, 2024. Association
  for Computing Machinery.

\bibitem{JSLS_TCSVT}
Qi Zang, Shuang Wang, Dong Zhao, Zhun Zhong, Biao Hou, and Licheng Jiao.
\newblock Joint style and layout synthesizing: Toward generalizable remote
  sensing semantic segmentation.
\newblock {\em IEEE Transactions on Circuits and Systems for Video Technology},
  pages 1--1, 2024.

\bibitem{zang2024ChangeDiff}
Qi Zang, Jiayi Yang, Shuang Wang, Dong Zhao, Wenjun Yi, and Zhun Zhong.
\newblock Changediff: A multi-temporal change detection data generator with
  flexible text prompts via diffusion model, 2024.

\bibitem{zhang2024improving}
Haojie Zhang, Yongyi Su, Xun Xu, and Kui Jia.
\newblock Improving the generalization of segmentation foundation model under
  distribution shift via weakly supervised adaptation.
\newblock In {\em Proceedings of the IEEE/CVF Conference on Computer Vision and
  Pattern Recognition}, pages 23385--23395, 2024.

\bibitem{zhang2023prompt}
Renrui Zhang, Xiangfei Hu, Bohao Li, Siyuan Huang, Hanqiu Deng, Yu Qiao, Peng
  Gao, and Hongsheng Li.
\newblock Prompt, generate, then cache: Cascade of foundation models makes
  strong few-shot learners.
\newblock In {\em Proceedings of the IEEE/CVF Conference on Computer Vision and
  Pattern Recognition}, pages 15211--15222, 2023.

\bibitem{Zhao_2024_CVPR_SND}
Dong Zhao, Shuang Wang, Qi Zang, Licheng Jiao, Nicu Sebe, and Zhun Zhong.
\newblock Stable neighbor denoising for source-free domain adaptive
  segmentation.
\newblock In {\em Proceedings of the IEEE/CVF Conference on Computer Vision and
  Pattern Recognition (CVPR)}, pages 23416--23427, June 2024.

\bibitem{Zhao_2023_CVPR}
Dong Zhao, Shuang Wang, Qi Zang, Dou Quan, Xiutiao Ye, and Licheng Jiao.
\newblock Towards better stability and adaptability: Improve online
  self-training for model adaptation in semantic segmentation.
\newblock In {\em Proceedings of the IEEE/CVF Conference on Computer Vision and
  Pattern Recognition (CVPR)}, pages 11733--11743, June 2023.

\bibitem{Zhao_2023_ICCV}
Dong Zhao, Shuang Wang, Qi Zang, Dou Quan, Xiutiao Ye, Rui Yang, and Licheng
  Jiao.
\newblock Learning pseudo-relations for cross-domain semantic segmentation.
\newblock In {\em Proceedings of the IEEE/CVF International Conference on
  Computer Vision (ICCV)}, pages 19191--19203, October 2023.

\bibitem{zhao2023semantic}
Dong Zhao, Ruizhi Yang, Shuang Wang, Qi Zang, Yang Hu, Licheng Jiao, Nicu Sebe,
  and Zhun Zhong.
\newblock Semantic connectivity-driven pseudo-labeling for cross-domain
  segmentation, 2023.

\bibitem{zhao2022style}
Yuyang Zhao, Zhun Zhong, Na Zhao, Nicu Sebe, and Gim~Hee Lee.
\newblock Style-hallucinated dual consistency learning for domain generalized
  semantic segmentation.
\newblock In {\em European conference on computer vision}, pages 535--552.
  Springer, 2022.

\bibitem{zhong2022adversarial}
Zhun Zhong, Yuyang Zhao, Gim~Hee Lee, and Nicu Sebe.
\newblock Adversarial style augmentation for domain generalized urban-scene
  segmentation.
\newblock {\em Advances in neural information processing systems}, 35:338--350,
  2022.

\bibitem{zhou2022conditional}
Kaiyang Zhou, Jingkang Yang, Chen~Change Loy, and Ziwei Liu.
\newblock Conditional prompt learning for vision-language models.
\newblock In {\em Proceedings of the IEEE/CVF conference on computer vision and
  pattern recognition}, pages 16816--16825, 2022.

\end{thebibliography}
}

\end{document}